\documentclass{article}

\usepackage{arxiv}

\usepackage[utf8]{inputenc}
\usepackage[T1]{fontenc}
\usepackage{hyperref}
\usepackage{url}
\usepackage{booktabs}
\usepackage{amsfonts}
\usepackage{nicefrac}
\usepackage[expansion=false]{microtype}
\usepackage{graphicx}
\usepackage{float}
\usepackage[section]{placeins}
\usepackage{amsmath,amssymb,amsthm}
\usepackage{multirow}
\usepackage{array}
\usepackage[dvipsnames]{xcolor}
\usepackage{caption}
\usepackage{enumitem}
\usepackage{subcaption}
\usepackage{cleveref}
\usepackage{setspace}
\usepackage{natbib}
\usepackage{doi}
\usepackage{adjustbox}
\usepackage{pifont}
\usepackage{tabularx}
\usepackage{mathrsfs}
\usepackage{bm}
\usepackage{colortbl}
\usepackage{textcomp}

\usepackage{algorithm}
\usepackage{algpseudocode}

\usepackage{tikz}
\usetikzlibrary{shapes.geometric, positioning, calc, fit}
\tikzset{myarrow/.style={-stealth, thick}}

\usepackage{siunitx}
\title{Force Sensor-Based Picking Pattern Analysis and Two-Jaw Gripper Design for Cherry Tomato Robotic Harvesting in Greenhouse}

\author{
  Rajmeet Singh Bhourji$^{*}$ \\
  Mohamed bin Zayed University of Artificial Intelligence\\
  Abu Dhabi, UAE \\
  Khalifa University Center for Autonomous Robotic Systems (KUCARS)\\
  Khalifa University, Abu Dhabi, UAE \\
  \texttt{rajmeet.bhourji@mbzuai.ac.ae}
  \And
  Manveen Kaur \\
  Department of Mechanical Engineering\\
  University of Windsor\\
  ON, Canada \\
  \texttt{Kaur79@uwindsor.ca}
  \And
  Yusuf Adamu \\
  Khalifa University Center for Autonomous Robotic Systems (KUCARS)\\
  Khalifa University, Abu Dhabi, UAE \\
  \texttt{yusuf.adamu@ku.ac.ae}
  \And
  Irfan Hussain \\
  Khalifa University Center for Autonomous Robotic Systems (KUCARS)\\
  Khalifa University, Abu Dhabi, UAE \\
  \texttt{irfan.hussain@ku.ac.ae}
  \And
  \vspace{-1em}
  \normalsize\textit{$^{*}$ Corresponding author.}
}

\hypersetup{
  pdftitle={Force Sensor-Based Picking Pattern Analysis and Two-Jaw Gripper Design for Cherry Tomato Robotic Harvesting in Greenhouse},
  pdfsubject={Agricultural Robotics, Robotic Harvesting, Force Sensing, Gripper Design, Greenhouse Automation},
  pdfauthor={Rajmeet Singh Bhourji, Manveen Kaur, Yusuf Adamu, Irfan Hussain},
  pdfkeywords={Cherry tomato harvesting, ATI force/torque sensor, Dynamixel servo gripper, UR5 manipulator, Husky A200, Mobile harvesting robot, Greenhouse automation},
  colorlinks=true,
  citecolor=blue,
  linkcolor=blue,
  urlcolor=blue
}

\begin{document}
\maketitle

% ── Abstract ──────────────────────────────────────────────────────────────────
\begin{abstract}
Robotic harvesting of cherry tomatoes in greenhouse environments remains challenging due to complex growth patterns, dense fruit clusters, delicate fruit skin, and the need for gentle yet effective detachment strategies. This study presents a comprehensive force-data-driven framework for autonomous cherry tomato harvesting that integrates six-axis force/torque sensing, a novel bio-inspired compliant gripper, and end-to-end deployment on a mobile robotic platform. A wearable ATI six-axis force/torque sensor was mounted on the index finger to capture the complete six-axis kinetic signatures during three elemental picking patterns gripping, twisting, and axial pulling across 50 manual trials conducted in both controlled laboratory and real greenhouse environments. The analysis revealed that gripping requires a peak normal force of $3.92 \pm 0.45$~N, twisting generates a dominant rotational torque of $19.20 \pm 2.10$~N$\cdot$mm about the pedicel axis with minimal truss disturbance, while pure axial pulling alone is insufficient for reliable detachment. These measurements directly informed the design of a novel fingertip localised Finray gripper in which the compliant Finray geometry is restricted to the fruit contact fingertip region (3D-printed in TPU 95A) while the supporting body is fabricated from rigid PLA, emulating the dual stiffness architecture of the human fingertip. The gripper is actuated by a Dynamixel XM430-W350 smart servomotor under closed-loop current based torque control, with the gripping force and $45^{\circ}$ wrist twist angle parameters directly transferred from the measured force envelopes. Finite element analysis confirmed a maximum von Mises stress of only 0.24~MPa under the calibrated 3.92~N gripping load, well below the bruising threshold of cherry tomato skin. The proposed gripper was integrated onto a UR5 manipulator mounted on a Clearpath Husky~A200 mobile base and validated first in a high fidelity NVIDIA Isaac Sim environment and subsequently through 50 fully autonomous real-world harvesting trials at the SILAL commercial greenhouse facility in Abu Dhabi, UAE. The integrated \textbf{hybrid grip--twist--pull} algorithm achieved a picking success rate of \textbf{85.0\%}, a fruit damage rate of only \textbf{2.0\%}, and an average per-fruit cycle time of \textbf{8.4~s}, substantially outperforming state-of-the-art benchmark end-effectors including vacuum (66.3\%), rotating (70.1\%), clamping (76.0\%), and suction cutting (52.4\%) grippers. This work establishes a practical framework that bridges high fidelity six-axis force analysis with bio-inspired servo-driven end-effector design for damage free autonomous greenhouse cherry tomatoes harvesting.
\end{abstract}

\keywords{Cherry tomato harvesting \and ATI force/torque sensor \and Dynamixel servo gripper \and UR5 manipulator \and Husky A200 \and Mobile harvesting robot \and Greenhouse automation}

% ═════════════════════════════════════════════════════════════════════════════
% 1. Introduction
% ═════════════════════════════════════════════════════════════════════════════
\section{Introduction}
\label{sec:introduction}

Tomatoes rank among the most widely cultivated and consumed vegetable crops
worldwide, with a global planting area exceeding 5.05 million hectares and an
annual production surpassing 182 million metric tonnes \citep{gao2022development}.
China alone accounts for a substantial share of this output, with a cultivation
area exceeding 1.1 million hectares and an annual yield of approximately 65.15
million tonnes \citep{huajing2021}, of which greenhouse grown tomatoes occupy
roughly 642,000 hectares \citep{ymt2022}. Despite this enormous scale of
production, tomato harvesting remains overwhelmingly dependent on manual labour,
with the picking stage alone accounting for more than 50\% of the total labour
demand across the entire planting and production cycle \citep{ren2020review}.
This heavy reliance on manual picking leads to persistently high labour costs,
low operational efficiency, and physically demanding workloads for farm workers.
These challenges are further compounded by growing labour shortages and rising
wage pressures affecting agricultural industries globally, threatening the
long term sustainability of tomato production \citep{ishii2021tomato,
kootstra2021selective}. In response, standardised greenhouse cultivation has
been promoted worldwide in recent years, creating controlled and structured
growing environments that provide favourable conditions for the introduction of
robotic harvesting systems \citep{gao2024lacta}. The development of
tomato harvesting robots is therefore recognised as a critical step toward
addressing the current labour crisis and improving the efficiency, consistency,
and scalability of greenhouse tomato production \citep{rapado2023development}.

The end effector is a key component of the harvesting robot, as its
design and performance directly determine the success rate, fruit
quality, and overall operational efficiency of the entire harvesting
system \citep{kultongkham2021design, zhang2018comparative}. In robotic
harvesting, one of the primary challenges is how to reliably detach
the fruit from the plant, where picking patterns and stem branch
characteristics significantly influence the required detachment forces
\citep{li2019anthropometric, hu2024effects, li2019factors}. The study
of picking patterns is therefore critical, as it directly affects
final harvesting performance and provides essential guidance for
end-effector design \citep{fan2021three}. Despite significant research
advances, current picking end effectors for cherry tomatoes continue
to face several persistent challenges, among which the lack of precise
force control during grasping and detachment remains the most critical.
Excessive gripping forces applied by rigid or poorly regulated end
effectors are a primary cause of fruit bruising, skin rupture, and
internal tissue damage, significantly reducing post-harvest shelf life
and commercial value \citep{zhang2018comparative, li2019factors}. This
problem is particularly acute for cherry tomatoes, whose small size
(typically 25--35~mm diameter), thin skin, and soft flesh make them
highly susceptible to mechanical damage even under moderate contact
pressures \citep{liu2020experimental}. Beyond fruit damage, existing
end effectors also suffer from large physical dimensions causing
interference with densely clustered fruits on the truss, excessive
disturbance distances that displace neighbouring fruits during picking,
and low picking success rates typically ranging from 52--76\%
\citep{gao2024picking, fujinaga2021development, xie2022optimization}.
The fundamental limitation underlying these issues is the absence of
quantitative force data to inform end-effector design: most existing
grippers are developed through empirical tuning or simplified force
assumptions rather than high fidelity measurements of the actual
forces and torques involved in different picking patterns during fruit
detachment \citep{gao2024picking, huang2021picking}. Without accurate
knowledge of the force thresholds associated with each detachment
strategy, it is difficult to design grippers that apply sufficient
force for reliable picking while remaining below the damage threshold
of the fruit. This gap between the required detachment forces and the
fruit damage limits represents a critical unresolved challenge in
cherry tomato harvesting robotics, and directly motivates the ATI
force/torque sensor based picking pattern analysis presented in this
study.

Understanding the biomechanics of fruit detachment has been a
recurring focus in harvesting robotics research, with several
investigations attempting to quantify the forces involved in manual
picking across different crop types. In the apple harvesting domain,
dynamic force analysis during hand picking revealed that detachment
performance varies significantly depending on both the picking
strategy employed and the specific fruit variety, highlighting the
need for crop specific force characterisation
\citep{davidson2016hand}. Subsequent work on apple detachment
mechanics demonstrated that certain robotic picking motions can
closely approximate the efficiency of human hand picking when the
biomechanical parameters are properly matched
\citep{li2016characterizing}. For mushroom harvesting, a combined
sensing approach integrating fingertip pressure measurement with
inertial motion tracking enabled the comparison of four distinct
harvesting strategies based on their force and motion signatures
\citep{huang2021picking}. Most relevant to the present study, a
systematic evaluation of cherry tomato detachment strategies
identified four candidate picking patterns and ranked twisting as
the most promising approach for robotic implementation, based on
fingertip force measurements collected with thin-film resistive
sensors \citep{gao2024picking}. Although these prior investigations
have collectively advanced understanding of crop detachment
mechanics, they all share a fundamental sensing limitation: the
measurement instruments employed, whether thin-film resistive
elements or discrete pressure transducers, are restricted to
capturing normal contact force at individual points. Such sensors
are inherently blind to lateral shear loading, pedicel bending
moments, and the torsional torque that governs twisting based
detachment, all of which are essential for understanding the
complete stress state at the fruit pedicel interface during
picking. Consequently, the full six-axis force and torque envelope
needed to establish quantitative design specifications for a
servo-driven gripper, including gripping force limits, rotation
torque thresholds, and pull force boundaries, cannot be derived
from existing datasets. This study directly addresses this
measurement gap by deploying an ATI six-axis force/torque
sensor capable of simultaneously recording all three force
components ($F_x$, $F_y$, $F_z$) and three torque components
($T_x$, $T_y$, $T_z$) throughout the cherry tomato picking
process, thereby providing the high resolution biomechanical
foundation required for the data-driven design of the proposed
gripper for cherry tomatoes harvesting.

A wide range of end-effector mechanisms have been proposed for
tomato harvesting over the past decade, each adopting a different
detachment strategy with varying degrees of success. Vacuum-based
approaches that combine suction with a cutting action have been
tested on large tomato varieties, but field trials reported picking
success rates as low as 52.4\%, primarily due to obstruction by
surrounding foliage and stems that prevent the suction cup from
achieving reliable contact with the fruit surface
\citep{fujinaga2021development}. Rotational plucking mechanisms,
which grip the fruit and twist it free from the stem, have shown
marginally better performance at approximately 60\% success, though
these systems are prone to disturbance during the twisting motion
and require relatively long cycle times per fruit
\citep{yaguchi2016development}. Shearing-based systems that sever
the peduncle with a blade have been applied to cherry tomato bunch
harvesting, but this approach is inherently unsuitable for
selective single-fruit picking and often requires multiple attempts
per successful harvest \citep{feng2018design}. Clamping grippers
designed for individual cherry tomato picking have achieved higher
success rates of up to 76\%, yet the rigid jaw contact and
uncontrolled clamping forces result in unacceptably high fruit
damage rates and large disturbance distances to adjacent fruits on
the truss, making them impractical for continuous picking
operations \citep{xie2022optimization}. Suction-shearing hybrid
mechanisms developed for large tomatoes similarly suffer from
bulky form factors and poor adaptability to the closely spaced
fruit arrangements typical of cherry tomato trusses
\citep{jun2021towards}. More recently, pneumatically actuated
vacuum and rotating end effectors designed specifically based on
cherry tomato picking pattern analysis achieved success rates of
66.3\% and 70.1\% respectively, representing the current
state-of-the-art, though detachment failure and collision with
neighbouring fruits remained the dominant failure modes
\citep{gao2024picking}. 

Across all these approaches, a common set
of shortcomings persists: oversized mechanisms that cannot navigate
the tight spacing between fruits on a truss, open-loop or
pneumatic actuation that lacks the fine-grained force regulation
needed to prevent bruising, substantial disturbance to the plant
canopy during picking that disrupts subsequent harvesting attempts,
and overall success rates that remain far below the reliability
threshold required for commercial deployment. These limitations
collectively underscore the need for a compact, force-controlled
gripper whose design parameters are derived directly from
quantitative detachment force measurements rather than empirical
approximation.

\subsection{Research Gaps}

Despite notable progress in tomato harvesting robotics, a critical
review of the existing literature reveals three fundamental gaps
that collectively hinder the development of reliable and
damage-free cherry tomato picking systems:

\begin{enumerate}[label=(\arabic*)]
    \item \textbf{Inadequate force sensing in picking pattern
    studies:} All existing investigations of cherry tomato picking
    biomechanics have relied on thin-film resistive pressure sensors
    or discrete pressure transducer arrays that capture only normal
    contact force at isolated fingertip locations. These instruments
    are inherently incapable of measuring lateral shear forces,
    pedicel bending moments, or the torsional torque that governs
    twisting-based detachment. As a result, the complete six-axis
    force/torque envelope required to define safe gripping thresholds
    and torque limits for gripper design remains unavailable from any
    prior study.

    \item \textbf{Lack of compliant, force-regulated end effectors:}
    The majority of reported cherry tomato end effectors employ
    either rigid jaw mechanisms with uncontrolled clamping forces or
    pneumatic actuation systems that offer only binary open/close
    states with no continuous force modulation. Neither approach can
    replicate the adaptive, conformable grasp of the human hand that
    distributes contact pressure evenly across the fruit surface to
    prevent localised bruising. Bio-inspired compliant structures
    such as FinRay-based fingers, which naturally conform around
    curved surfaces under load, have not yet been explored for cherry
    tomato harvesting despite their demonstrated effectiveness in
    delicate object manipulation.

    \item \textbf{Absence of an integrated force-to-gripper design
    pipeline:} No existing work establishes a direct, quantitative
    link from high-fidelity picking force measurements through to
    servo-driven gripper design, simulation validation, and
    real-world greenhouse deployment on a mobile manipulation
    platform. Current end effectors are designed through empirical
    trial-and-error rather than being parametrically derived from
    measured detachment force data, resulting in grippers whose
    operating forces are neither optimised for reliable detachment
    nor constrained below the fruit damage threshold.
\end{enumerate}

The present study is specifically designed to address all three of
these gaps through the integration of ATI six-axis
force/torque sensing, TPU FinRay-inspired compliant gripper fingers,
Dynamixel servo motor closed-loop force control, and end-to-end
validation on a Husky A200--UR5 mobile harvesting platform.

The main contributions of this work are as follows:

\begin{itemize}
    \item \textbf{ATI force/torque measurement for picking
    pattern characterisation:} An ATI six-axis force/torque
    sensor is mounted on the operator's thumb to capture the complete
    force and torsional torque profiles during three fundamental
    cherry tomato picking patterns (grasping, twisting, and pulling).
    Data collection is conducted across both controlled laboratory
    conditions and real greenhouse environments at the SILAL facility
    in Abu Dhabi, ensuring that the measured force envelopes reflect
    the variability encountered in practical harvesting scenarios.

    \item \textbf{Force-data-driven bio-inspired gripper design:} The
    measured force and torque profiles are used to establish
    quantitative design specifications, including maximum gripping
    force limits, torsional torque thresholds, and pull force
    boundaries, for a novel two-finger gripper. The gripper fingers
    are fabricated from thermoplastic polyurethane (TPU) using a
    FinRay-inspired compliant structure, enabling the two fingers to
    conform around the fruit surface in a manner that replicates the
    natural opposition of the human thumb and index finger during
    manual picking. This compliant contact distribution minimises
    localised stress concentrations on the fruit skin, allowing
    secure grasping without bruising or surface damage.

    \item \textbf{Dynamixel servo motor actuation with closed-loop
    force control:} The gripper is actuated by Dynamixel XM430-W350
    servo motors operating in current-based torque control mode,
    enabling direct mapping from the ATI-measured picking force
    profiles to real-time motor commands. This closed-loop approach
    ensures that gripping forces remain within the safe operating
    envelope derived from the sensor measurements, eliminating the
    reliance on empirical force tuning that limits existing
    pneumatic end effectors.

    \item \textbf{Mobile harvesting platform integration and dual
    validation:} The proposed gripper is integrated onto a UR5
    robotic manipulator mounted on a Clearpath Husky A200 mobile
    base, forming a complete autonomous harvesting platform. System
    performance is validated through both simulation and
    real-world picking trials at the SILAL greenhouse, demonstrating
    end-to-end functionality from force-informed design through to
    autonomous field deployment.
\end{itemize}

% ═════════════════════════════════════════════════════════════════════════════
% 2. Materials And Methods
% ═════════════════════════════════════════════════════════════════════════════
\section{Materials and Methods}
\label{sec:materials}

\subsection{Experimental Site and Fruit Characterisation}

\begin{figure}[!ht]
    \centering
    \includegraphics[width=0.6\linewidth]{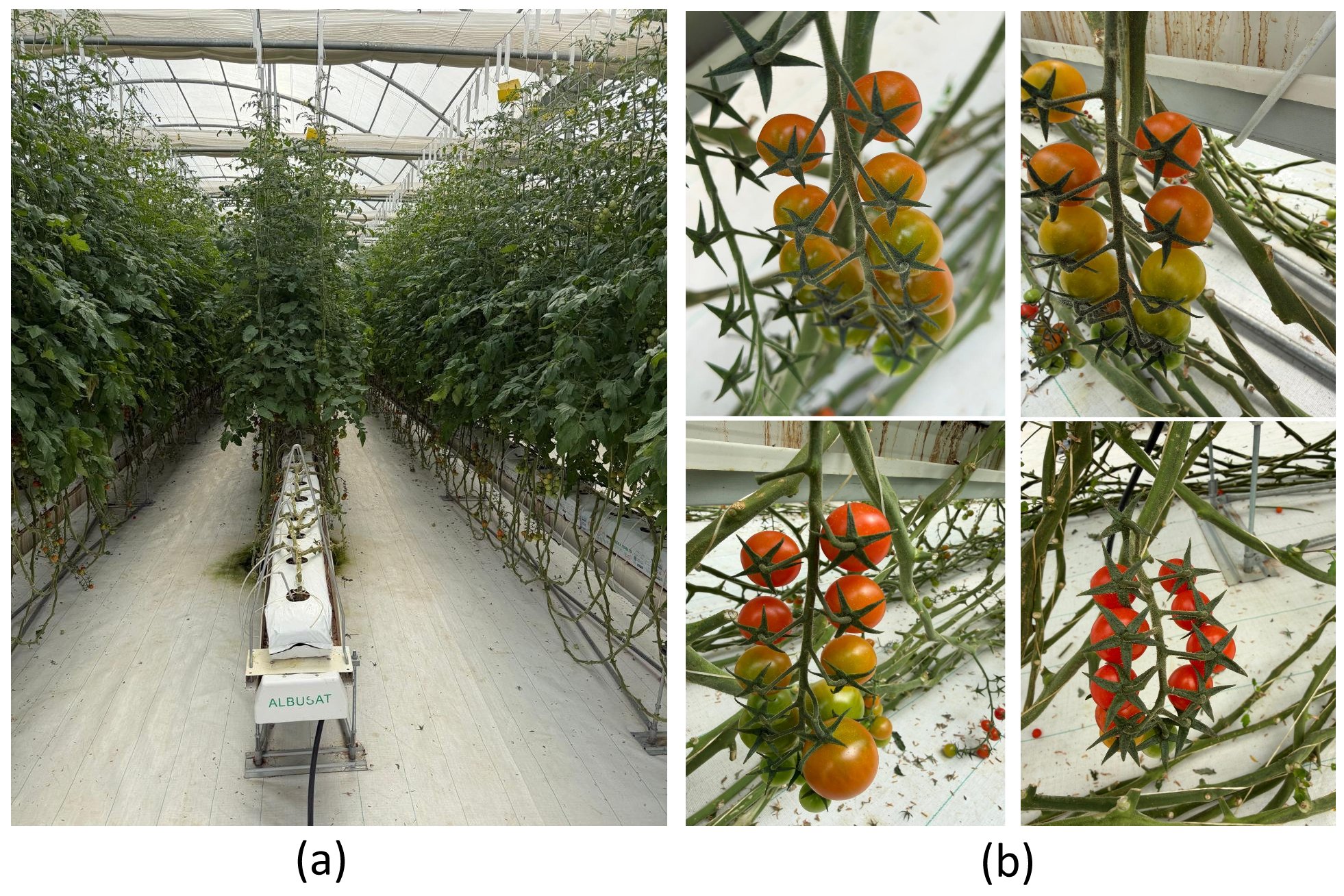}    
    \caption{Typical planting pattern of cherry tomatoes in the SILAL 
    commercial greenhouse, Abu Dhabi, UAE. (a) Overview of high-wire 
    trellising system. (b) Close-up of cherry tomato truss with ripe 
    fruits. }
    \label{fig:fig1}
\end{figure}

Experiments were conducted at the SILAL greenhouse facility in Abu Dhabi, UAE, a commercial-scale controlled-environment agriculture operation as depicted in Fig.~\ref{fig:fig1}(a). The cherry tomatoes are grown on high-wire trellising systems in standardised rows, with plants spaced approximately 500~mm within rows and 1600~mm between rows. Fig.~\ref{fig:fig1}(b) shows the growing pattern of the cherry tomatoes. 

Table~\ref{tab:table1} presents the physical parameters of the cherry tomato samples measured at the SILAL greenhouse facility.

\begin{table}[!ht]
\centering
\caption{Means and standard deviations (SD) of physical parameters of cherry tomato fruit samples measured at SILAL greenhouse.}
\label{tab:table1}
\small
\renewcommand{\arraystretch}{1.2}
\begin{tabular}{llcc}
\toprule
\textbf{Category} & \textbf{Units} & \textbf{Mean} & \textbf{SD} \\
\midrule
Fruit diameter       & (mm)  & 30.25  & 1.30 \\
Fruit height       & (mm)  & 26.10  & 1.12 \\
Truss weight     & (g)    & 168.06  & 15.50 \\
Pedicel length   & (mm)    & 43.52  & 6.44 \\
Fruits number      & (numbers)        & 15  & 3 \\

\bottomrule
\end{tabular}
\end{table}

\subsection{Picking patterns}

\begin{figure}[!ht]
    \centering
    \includegraphics[width=\linewidth]{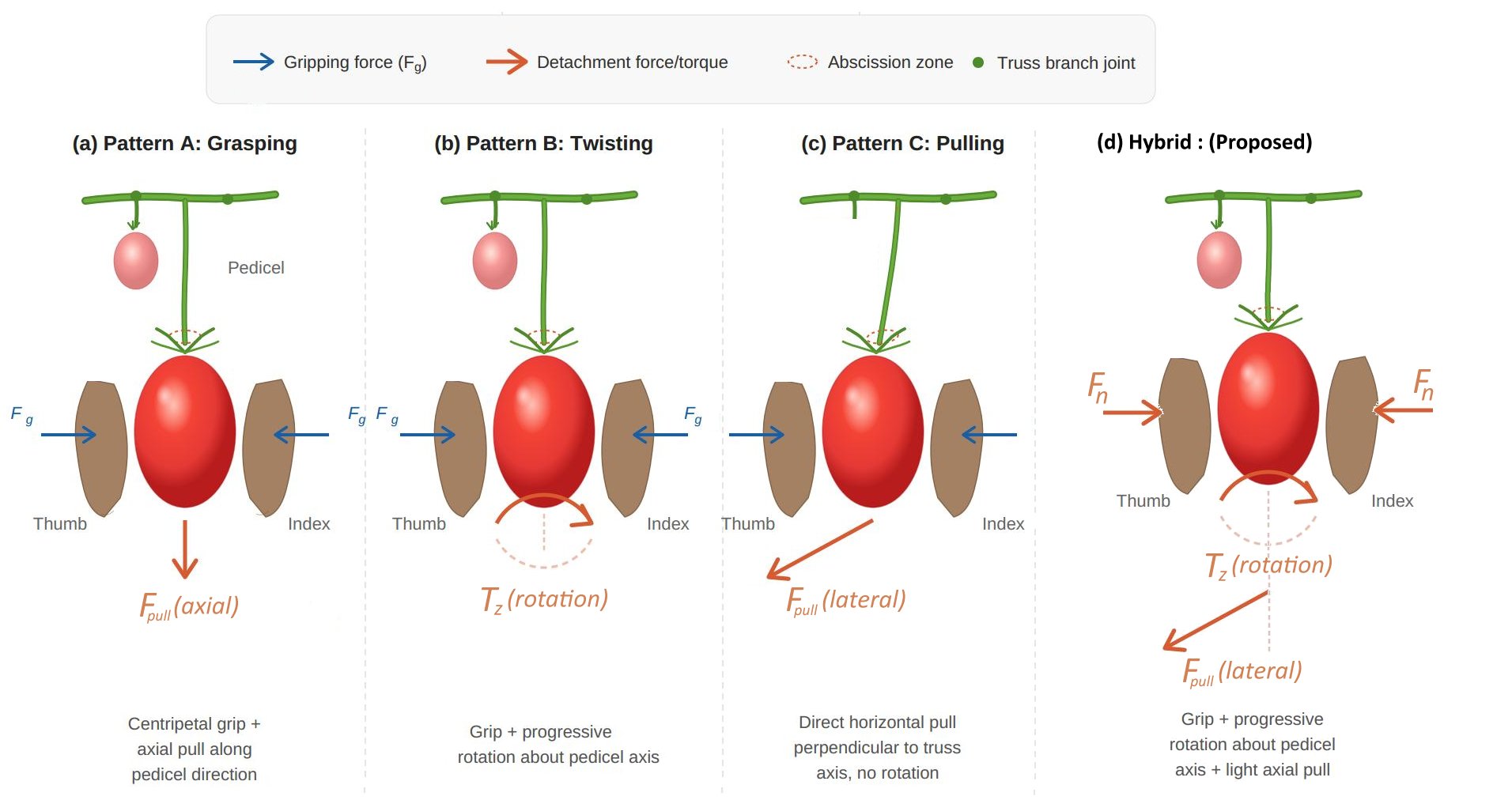}
    \caption{Cherry tomato picking patterns derived from manual
    harvesting observation at the SILAL greenhouse facility.
    (a)~Pattern~A: Grasping centripetal grip with axial pull
    ($F_{pull}$) along the pedicel direction, causing tensile pedicel
    failure. (b)~Pattern~B: Twisting grip with progressive
    rotation ($T_z$) about the pedicel axis, causing torsional shear
    failure. (c)~Pattern~C: Pulling direct horizontal pull
    ($F_{pull}$) perpendicular to the truss axis, causing tensile
    and bending failure. (d)~Hybrid (Proposed) the proposed
    robotic picking strategy combining progressive rotation ($T_z$)
    with light lateral pull ($F_{pull}$), applying normal forces
    ($F_n$) axially to secure the fruit while minimising detachment
    force and truss disturbance. Blue arrows denote gripping forces
    ($F_g$); orange arrows denote detachment forces and torques.
    Dashed ellipses indicate the abscission zone at the
    pedicel calyx junction.}
    \label{fig:fig2}
\end{figure}

To establish a biomechanically grounded basis for gripper design,
this study began by observing experienced greenhouse workers at the
SILAL facility during routine manual cherry tomato harvesting.
Analysis of the observed hand movements revealed that all manual
picking actions can be decomposed into three elemental motion
primitives: grasping with a direct axial pull, rotational twisting
about the pedicel axis, and lateral pulling perpendicular to the
truss. These three primitives were formalised into distinct picking
patterns. In addition to the three baseline patterns, a fourth
hybrid pattern combining axial normal, twisting and lateral pulling is proposed
as the intended robotic picking strategy. All four patterns are
illustrated in Fig.~\ref{fig:fig2}(a-d).

\begin{itemize}
    \item \textbf{Grasping (Pattern A) Fig.~\ref{fig:fig2}(a):} The thumb and index finger
    enclose the fruit at the equatorial plane, establishing a stable
    centripetal grip, after which a steady axial force is applied
    downward along the pedicel direction to separate the fruit from
    the truss through tensile failure at the abscission zone.

    \item \textbf{Twisting (Pattern B) Fig.~\ref{fig:fig2}(b):} The thumb and index finger
    grip the fruit with moderate contact force and progressively
    rotate it about the pedicel longitudinal axis. Detachment occurs
    through torsional shear failure of the pedicel fibres at the
    abscission zone, typically requiring two to four rotation cycles
    depending on pedicel maturity and diameter.

    \item \textbf{Pulling (Pattern C) Fig.~\ref{fig:fig2}(c):} A direct lateral force is
    applied horizontally, perpendicular to the truss axis, without
    any rotational component. Detachment occurs through combined
    tensile and bending failure at the pedicel base. This is the
    mechanically simplest motion but generates the highest peak
    forces and the largest disturbance to adjacent fruits on the
    truss.

    \item \textbf{Hybrid (Pattern D Proposed) Fig.~\ref{fig:fig2}(d):} The proposed
    robotic picking strategy combines elements of Patterns A, B and C.
    The thumb and index finger grip (normal force) the fruit and apply progressive
    rotation about the pedicel axis while simultaneously exerting a
    light lateral pull. This hybrid approach leverages the low-torque
    torsional weakening from twisting to reduce the lateral force
    needed for final separation, thereby minimising both the peak
    detachment force and the disturbance distance to neighbouring
    fruits. The normal forces ($F_n$) applied by each finger are
    oriented axially inward toward the fruit centre, providing
    stable grip without excessive surface pressure.
\end{itemize}

\subsection{Force/Torque Measurement System}

\begin{figure}[!ht]
    \centering
    \includegraphics[width=0.6\linewidth]{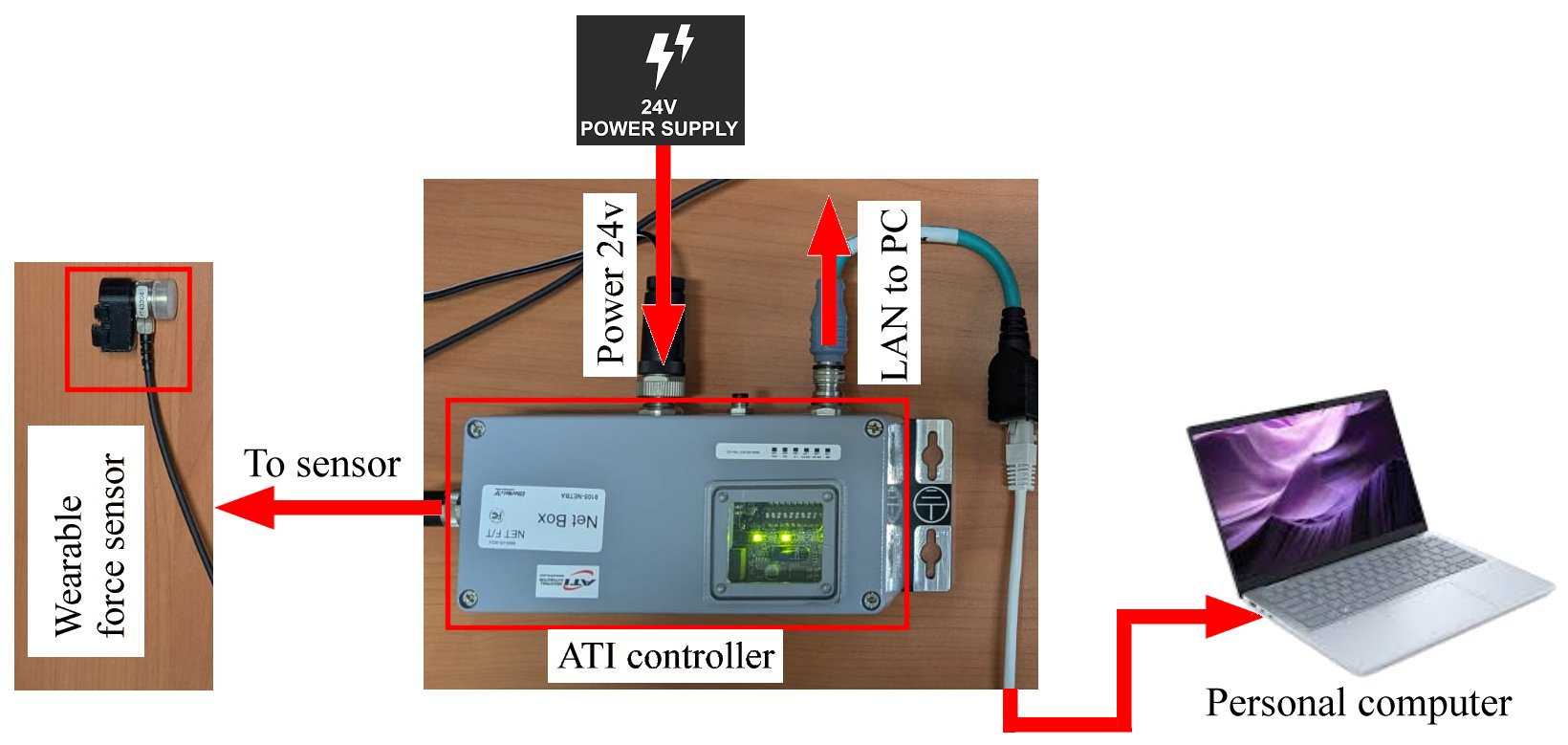}    
    \caption{Measurement setup of the ATI force sensor-based data acquisition system. The wearable ATI force sensor mounted on the finger is connected to the ATI Net Box controller, which receives a 24 V DC power supply and transmits the acquired force and torque data to a personal computer via a LAN (Ethernet) connection for real-time monitoring, processing, and storage.}
    \label{fig:fig3}
\end{figure}

Comparing the applied force and torsional torque variations throughout the picking process serves as an effective approach for evaluating and differentiating between various picking patterns. In this study, a dynamic measurement system was designed and developed to capture key parameters, including applied force and torsional torque, during the cherry tomato harvesting operation. As illustrated in Fig.~\ref{fig:fig3}. 

The ATI FT43304 six-axis force/torque sensor (ATI Industrial Automation, Apex, NC, USA) was selected for the dynamic force measurement experiments. The ATI sensor measures forces and torques along three orthogonal axes ($F_x$, $F_y$, $F_z$, $T_x$, $T_y$, $T_z$). Table~\ref{tab:table2} summarises the sensor specifications. 

The sensor communicates via a Net F/T interface providing data acquisition 
rates up to 7000~Hz, enabling high-temporal-resolution capture of the dynamic 
picking process. The sensor weighs 38.7~g with a diameter of 43~mm and height 
of 11.5~mm, making it suitable for finger-mounted applications without 
significantly altering the natural picking motion. The sensor measures forces 
in the $F_x$ and $F_y$ axes up to $\pm$145~N with a resolution of 
$\frac{1}{16}$~N, and up to $\pm$290~N in the $F_z$ axis with a resolution 
of $\frac{1}{8}$~N, while torque measurements across all axes ($T_x$, $T_y$, 
$T_z$) range up to $\pm$5~N$\cdot$m with a resolution of 
$\frac{1}{376}$~N$\cdot$m. The overload limits of $\pm$300~N for 
$F_x$/$F_y$, $\pm$900~N for $F_z$, and $\pm$9~N$\cdot$m for all torque 
axes ensure robust performance under the peak forces encountered during the 
picking trials. Furthermore, the stiffness values of 
$1.1 \times 10^{7}$~N/m for $F_x$/$F_y$, $1.7 \times 10^{7}$~N/m for 
$F_z$, and $1.4 \times 10^{4}$~N$\cdot$m/rad for torque axes confirm the 
rigid mechanical behaviour of the sensor, minimising deflection induced 
measurement errors during high speed dynamic trials.

\begin{table}[!ht]
\centering
\caption{ATI force/torque sensor specifications.}
\label{tab:table2}
\small
\renewcommand{\arraystretch}{1.2}
\begin{tabular}{lccc}
\toprule
\textbf{Parameter} & \textbf{Range} & \textbf{Resolution} & \textbf{Overload Limit} \\
\midrule
$F_x$, $F_y$   & $\pm$145~N         & 1/16~N              & $\pm$300~N \\
$F_z$           & $\pm$290~N         & 1/8~N               & $\pm$900~N \\
$T_x$, $T_y$   & $\pm$5~N$\cdot$m   & 1/376~N$\cdot$m     & $\pm$9~N$\cdot$m \\
$T_z$           & $\pm$5~N$\cdot$m   & 1/376~N$\cdot$m     & $\pm$9~N$\cdot$m \\
\midrule
\multicolumn{4}{l}{\textit{Physical Properties}} \\
\midrule
Diameter        & \multicolumn{3}{c}{43~mm} \\
Height          & \multicolumn{3}{c}{11.5~mm} \\
Weight          & \multicolumn{3}{c}{38.7~g (0.0387~kg)} \\
\midrule
\multicolumn{4}{l}{\textit{Stiffness}} \\
\midrule
$F_x$, $F_y$   & \multicolumn{3}{c}{$1.1 \times 10^7$~N/m} \\
$F_z$           & \multicolumn{3}{c}{$1.7 \times 10^7$~N/m} \\
Torque          & \multicolumn{3}{c}{$1.4 \times 10^4$~N$\cdot$m/rad} \\
\bottomrule
\end{tabular}
\end{table}

\subsubsection{Sensor mounting configuration}
\begin{figure}[!ht]
    \centering
    \includegraphics[width=0.6\linewidth]{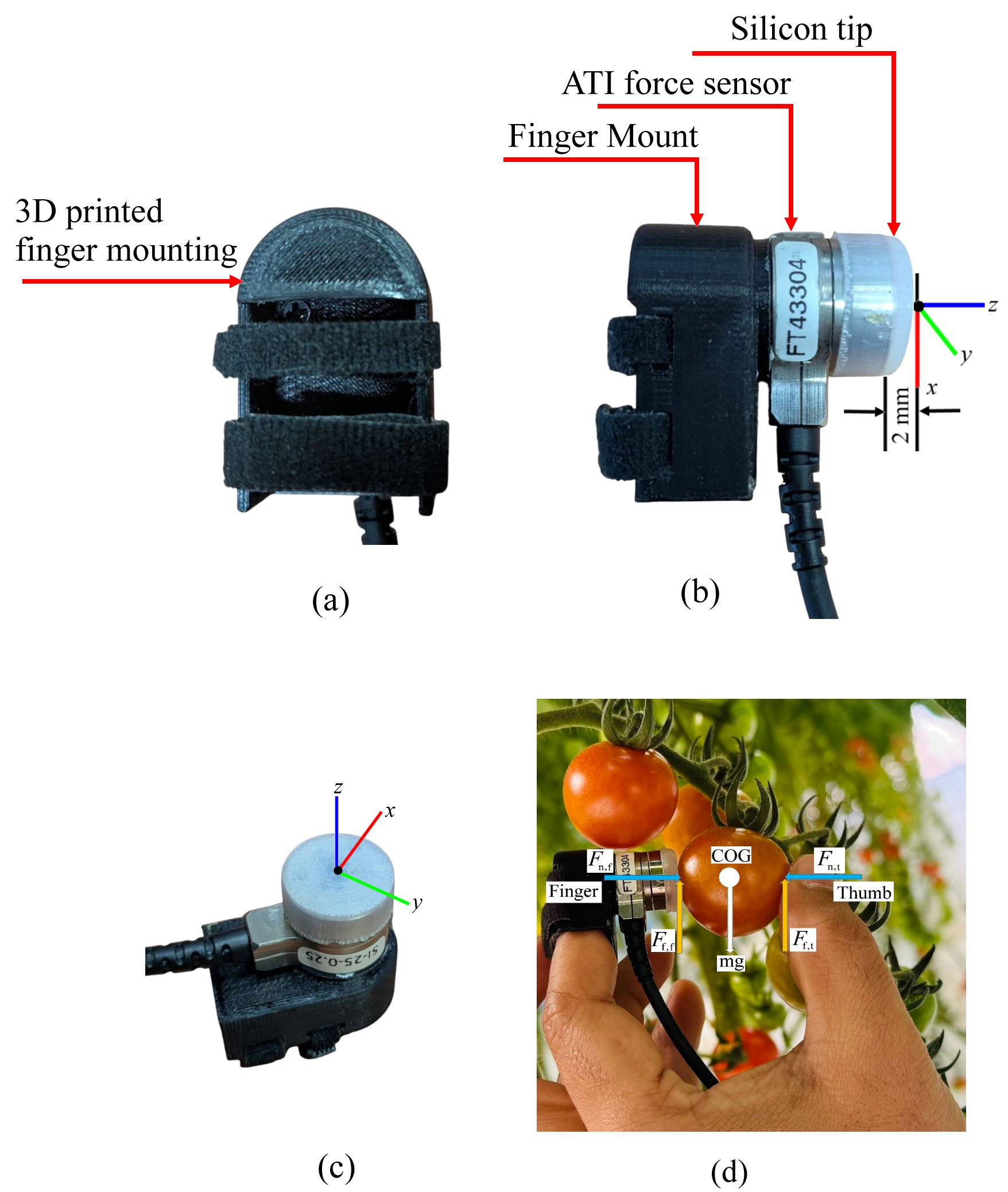}
    \caption{ATI force sensor finger-mounted assembly and force diagram 
    during tomato picking. (a) Front view of the 3D-printed finger 
    mounting bracket secured with adjustable straps for stable attachment 
    to the finger during picking trials. (b) Side view illustrating the 
    complete assembly comprising the 3D-printed finger mount, ATI force 
    sensor (FT43304), and silicon tip, with the three-axis coordinate 
    frame ($x$, $y$, $z$) indicating the sensing directions and a 2~mm 
    reference scale shown for dimensional context. (c) Isometric view of 
    the ATI force sensor integrated within the 3D-printed 
    mount, with the coordinate frame axes ($x$, $y$, $z$) denoting the 
    force and torque measurement directions. (d) Free-body diagram of the 
    applied forces during a two-finger tomato picking trial, showing the 
    normal forces ($F_\mathrm{n,f}$ and $F_\mathrm{n,t}$) and frictional 
    forces ($F_\mathrm{f,f}$ and $F_\mathrm{f,t}$) exerted by the index 
    finger and thumb on the tomato, respectively, along with the 
    gravitational force ($mg$) acting at the centre of gravity (COG) 
    of the tomato.}
    \label{fig:fig4}
\end{figure}

The ATI sensor was mounted on a custom 3D-printed adapter affixed to the index finger using an adjustable finger sleeve assembly as depicted in Fig.~\ref{fig:fig4}. The mounting design ensures that the sensor's Z-axis aligns with the primary direction of applied force during grasping, while the X and Y axes capture lateral and tangential forces. A thin silicone rubber pad (2~mm thickness) was placed between the sensor contact surface and the fruit to prevent bruising and ensure consistent contact conditions. The sensor was connected to the Net F/T data acquisition box via a shielded cable, with data streamed to a host PC running python for synchronised recording.

During the two finger tomato picking process, the forces acting on 
the tomato can be decomposed into normal forces and frictional forces 
exerted by the index finger and thumb, along with the gravitational 
force. Based on the free-body diagram illustrated in Fig.~\ref{fig:fig4}(d), 
the static force equilibrium conditions are expressed as follows.

The normal force equilibrium in the horizontal direction requires 
that the normal forces applied by the index finger and thumb are 
equal and opposite:
\begin{equation}
    F_{\mathrm{n,f}} = F_{\mathrm{n,t}}
    \label{eq:normal_eq}
\end{equation}

The vertical force equilibrium condition requires that the sum of 
the frictional forces generated by the index finger and thumb must 
support the weight of the tomato:
\begin{equation}
    F_{\mathrm{f,f}} + F_{\mathrm{f,t}} \geq mg
    \label{eq:friction_eq}
\end{equation}

\noindent where $F_{\mathrm{n,f}}$ and $F_{\mathrm{n,t}}$ are the 
normal forces exerted by the index finger and thumb, respectively; 
$F_{\mathrm{f,f}}$ and $F_{\mathrm{f,t}}$ are the corresponding 
frictional forces; $m$ is the mass of the tomato; and $g$ is the 
gravitational acceleration (9.81~m/s$^2$).

The frictional force at each contact point is governed by the 
Coulomb friction model:
\begin{equation}
    F_{\mathrm{f,f}} = \mu_{\mathrm{f}} \cdot F_{\mathrm{n,f}}
    \label{eq:coulomb_f}
\end{equation}
\begin{equation}
    F_{\mathrm{f,t}} = \mu_{\mathrm{t}} \cdot F_{\mathrm{n,t}}
    \label{eq:coulomb_t}
\end{equation}

\noindent where $\mu_{\mathrm{f}}$ and $\mu_{\mathrm{t}}$ are the 
coefficients of friction at the finger--tomato and thumb--tomato 
contact interfaces, respectively.

Substituting Eq.~(\ref{eq:normal_eq}) into 
Eqs.~(\ref{eq:coulomb_f}) and~(\ref{eq:coulomb_t}), and applying 
the vertical equilibrium condition of Eq.~(\ref{eq:friction_eq}), 
the minimum required normal force to prevent the tomato from 
slipping during the picking process is given by:
\begin{equation}
    F_{\mathrm{n,min}} = \frac{mg}{\mu_{\mathrm{f}} + \mu_{\mathrm{t}}}
    \label{eq:min_force}
\end{equation}

The total gripping force applied to the tomato during the picking 
process is therefore:
\begin{equation}
    F_{\mathrm{grip}} = F_{\mathrm{n,f}} + F_{\mathrm{n,t}} = 
    2F_{\mathrm{n,f}}
    \label{eq:grip_force}
\end{equation}

\subsection{Calibration and Data Acquisition}

Prior to each experimental trial, the ATI force/torque sensor 
was calibrated to ensure measurement accuracy and repeatability. 
The calibration procedure involved zeroing the sensor output in the 
unloaded condition to eliminate any bias or offset errors introduced 
by the weight of the finger mount assembly and silicon tip. This 
tare operation was performed through the ATI Net F/T software 
interface before each picking trial to ensure that only the contact 
forces generated during the picking process were recorded. The ATI sensor communicates with the personal computer via 
the Net F/T controller (Net Box) using a standard Ethernet (LAN) 
connection, powered by a 24~V DC supply. The Net F/T interface 
supports data acquisition rates of up to 7000~Hz, providing 
high-temporal-resolution force and torque measurements across all 
six axes ($F_x$, $F_y$, $F_z$, $T_x$, $T_y$, $T_z$) in real time. 
For the present study, the sampling rate was set to 1000~Hz to 
balance data resolution with storage requirements during dynamic 
picking trials.

The raw voltage signals from the sensor are conditioned and converted 
into calibrated force and torque values by the Net F/T controller 
using the sensor-specific calibration matrix provided by ATI 
Industrial Automation. The calibration matrix accounts for cross-axis 
coupling and sensitivity variations across the measurement range of 
$\pm$145~N for $F_x$/$F_y$, $\pm$290~N for $F_z$, and $\pm$5~N$\cdot$m 
for all torque axes ($T_x$, $T_y$, $T_z$), with resolutions of 
$\frac{1}{16}$~N, $\frac{1}{8}$~N, and $\frac{1}{376}$~N$\cdot$m, 
respectively.

The calibrated force and torque data were streamed in real time to 
the personal computer via the LAN interface and logged using the ATI 
Net F/T data acquisition software. The recorded data were subsequently 
exported in \texttt{.csv} format for post-processing and statistical 
analysis. 

\subsubsection{Experimental procedure}
\begin{figure}[!ht]
    \centering
    \includegraphics[width=\linewidth]{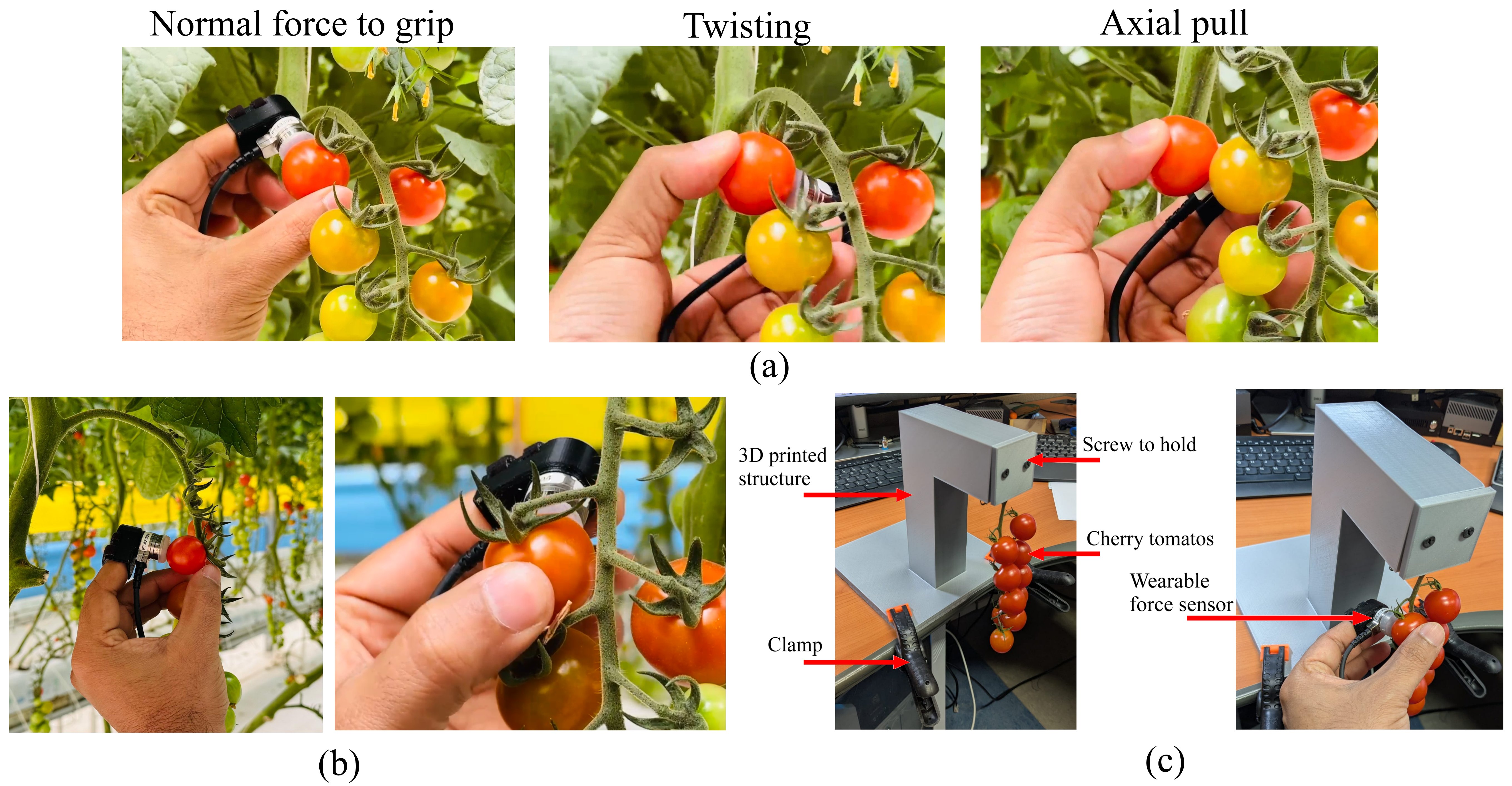}
    \caption{Experimental setup for cherry tomato picking force 
    measurement. (a) Three proposed hybrid picking motion patterns 
    applied during the harvesting trials: normal force to grip 
    (lateral pinch gripping), twisting (rotational detachment), 
    and axial pull (straight stem-pull detachment), each performed 
    with the ATI force sensor mounted on the index finger to capture 
    the corresponding six-axis force and torque profiles. (b) Sample 
    images of picking experimentation conducted in the 
    greenhouse environment, demonstrating the wearable ATI force 
    sensor deployment during actual cherry tomato harvesting on 
    the vine under real growing conditions. (c) Controlled 
    laboratory experimentation using a custom 3D-printed test bench, 
    comprising a rigid U-shaped 3D-printed structure with screw-based 
    clamping to fix the cherry tomato truss in a repeatable position, 
    enabling systematic force measurement with the wearable ATI force 
    sensor under standardised conditions.}
    \label{fig:fig5}
\end{figure}

Force measurement experiments were conducted at the SILAL greenhouse facility as well as In-Lab test bench as depicted in the Fig.~\ref{fig:fig5}. Each picking pattern was performed for 50 trials on ripe cherry tomatoes randomly selected from different trusses. All trials were performed by a single pre-trained operator wearing the ATI sensor-equipped finger sleeve. The operator was positioned approximately 40~cm from the plant. Before each trial, the ATI sensor was biased to zero. The following parameters were recorded during each trial: six-axis force/torque profiles ($F_x$, $F_y$, $F_z$, $T_x$, $T_y$, $T_z$) at 1000~Hz, picking duration, disturbance distance to the truss, and calyx retention status. After each trial, the detached fruit was weighed and measured. All data was synchronised and logged in\texttt{.csv} format.

\subsection{Novel Two-Jaw Gripper Design}

\begin{figure}[!ht]
    \centering
    \includegraphics[width=0.6\linewidth]{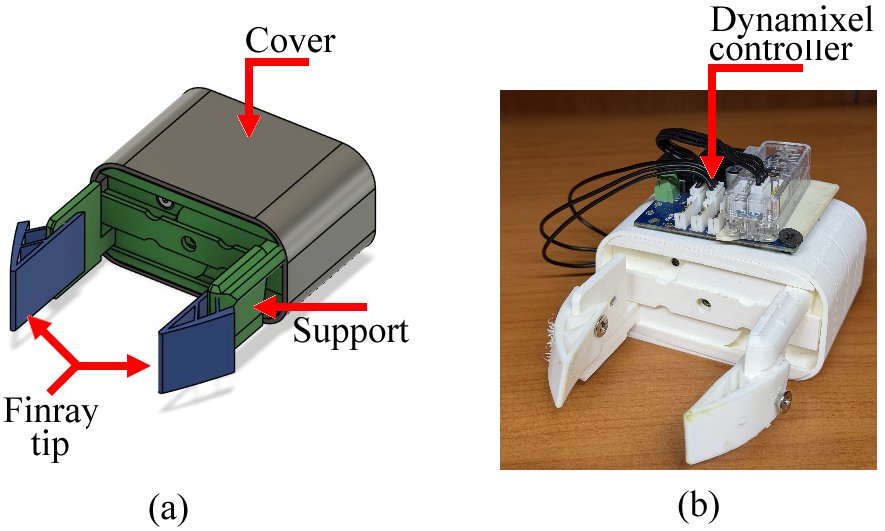}
    \caption{Proposed novel fingertip-localised Finray gripper for 
    damage-free cherry tomato harvesting. (a) CAD model of the 
    gripper assembly illustrating the major structural components, 
    including the outer protective cover, internal support frame, 
    and the Finray-inspired compliant fingertips fabricated from 
    flexible TPU 95A material for gentle conformal contact with the 
    tomato surface. (b) Fabricated prototype of the gripper, with 
    the rigid structural components 3D-printed in PLA material and 
    actuated by a Dynamixel XM430-W350 smart servomotor through 
    an integrated Dynamixel controller mounted on the top surface, 
    enabling closed-loop control of the gripping force and jaw 
    opening/closing motion.}
    \label{fig:fig6}
\end{figure}

Figure~\ref{fig:fig6} illustrates the proposed novel finger-tip 
Finray-based gripper developed for damage-free cherry tomato 
harvesting. Fig.~\ref{fig:fig6}(a) presents the CAD model of the 
gripper, showing the major structural components including the 
outer cover, internal support frame, and the Finray-inspired 
compliant tips, while Fig.~\ref{fig:fig6}(b) displays the 
corresponding fabricated prototype integrated with the actuation 
electronics. Unlike conventional Finray gripper designs reported 
in the literature, where the entire finger structure adopts the 
Finray geometry, the present work introduces a distinctive 
configuration in which the Finray-inspired compliant geometry is 
implemented \textit{only at the fingertip region}, specifically at 
the contact interface with the tomato surface. This novel 
localised application of the Finray principle enables the gripper 
to passively conform to the curvature of the tomato during contact, 
thereby distributing the contact pressure uniformly and preventing 
surface damage, bruising, or skin rupture during the picking 
process---a design strategy that, to the best of the authors' 
knowledge, has not been previously reported.

The Finray-inspired fingertip components are fabricated using 
flexible thermoplastic polyurethane (TPU 95A) material, which 
provides the necessary compliance, elastic deformation, and soft 
contact behaviour required for gentle interaction with the delicate 
tomato skin. The remaining structural components of the gripper, 
including the cover, support frame, and mounting brackets, are 
fabricated from polylactic acid (PLA) material using 
fused-deposition-modelling (FDM) 3D printing, providing the rigid 
structural backbone required to maintain the geometric integrity 
and mechanical stiffness of the gripper assembly. Actuation of the 
gripper jaws is achieved using a Dynamixel XM430-W350 smart 
servomotor (ROBOTIS, South Korea), which provides precise 
closed-loop control of both the gripping force and the jaw 
opening/closing motion. The integrated current-based torque 
feedback of the Dynamixel motor enables real-time monitoring and 
regulation of the applied gripping force, ensuring that the contact 
force remains within the safe handling threshold identified from the 
human picking trials, thereby preventing fruit damage while ensuring 
a secure grasp during harvesting.Table~\ref{tab:dynamixel_specs} summarises the key specifications.

\begin{table}[!ht]
\centering
\caption{Dynamixel XM430-W350 servo motor specifications.}
\label{tab:dynamixel_specs}
\small
\renewcommand{\arraystretch}{1.2}
\begin{tabular}{lc}
\toprule
\textbf{Parameter} & \textbf{Value} \\
\midrule
Stall Torque (12~V)     & 4.1~N$\cdot$m \\
No-Load Speed           & 46~RPM \\
Operating Voltage       & 10.0--14.8~V \\
Resolution              & 4096 pos/rev (0.088$^{\circ}$) \\
Communication           & TTL / RS-485 (Protocol 2.0) \\
Weight                  & 82~g \\
Dimensions              & 28.5 $\times$ 46.5 $\times$ 34~mm \\
Current-Based Control   & Yes (mode 0) \\
\bottomrule
\end{tabular}
\end{table}

\begin{figure}[!ht]
    \centering
    \includegraphics[width=0.6\linewidth]{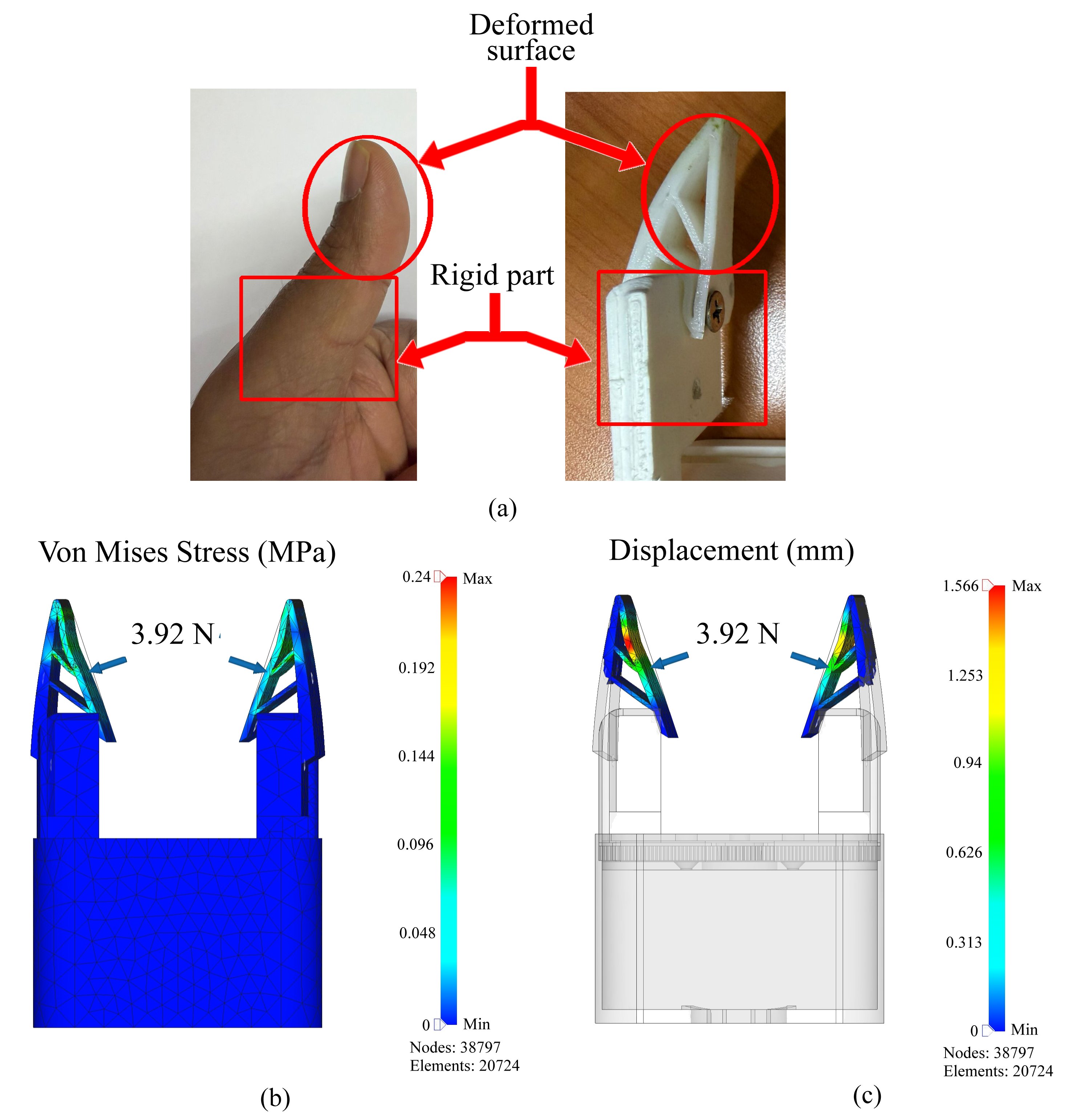}
    \caption{Bio-inspired design rationale and finite element 
    method (FEM) validation of the proposed fingertip-localised 
    Finray gripper for damage-free cherry tomato harvesting. 
    (a)~Conceptual analogy between the human fingertip and the 
    proposed gripper, illustrating the biomimetic dual-stiffness 
    configuration in which the soft, deformable distal fingertip 
    pad is replicated by the compliant TPU 95A Finray tip 
    (highlighted in red circle), while the rigid supporting 
    phalangeal structure is emulated by the PLA-printed rigid body 
    of the gripper (highlighted in red rectangle). (b)~Von Mises 
    stress distribution under an applied gripping force of 
    3.92~N derived from the average peak force measured across 
    50 manual cherry tomato picking trials using the wearable ATI 
   force sensor on the index finger showing a maximum 
    stress of 0.24~MPa localised within the Finray rib structure, 
    well below the yield strength of TPU 95A. (c)~Corresponding 
    total displacement distribution showing a maximum tip 
    deflection of 1.566~mm at the contact region, enabling 
    passive conformal adaptation to the tomato curvature.}
    \label{fig:fig7}
\end{figure}

Figure~\ref{fig:fig7} presents the conceptual bio-inspired design 
rationale and the corresponding finite element method (FEM) 
validation of the proposed fingertip-localised Finray gripper. As 
shown in Fig.~\ref{fig:fig7}(a), the proposed design is directly 
inspired by the natural biomechanical configuration of the human 
hand observed during cherry tomato picking. The human fingertip 
consists of a soft, deformable distal pad that conforms passively 
to the curved surface of the fruit, while the supporting phalangeal 
bone provides the structural rigidity required to transmit the 
gripping force. Mimicking this dual-stiffness principle, the 
proposed gripper integrates a compliant Finray-inspired tip 
fabricated from flexible TPU 95A material, which replicates the 
conformal, damage-free contact behaviour of the human fingertip 
pad, while the remaining rigid body of the gripper, 3D-printed in 
PLA, emulates the structural role of the underlying phalangeal 
skeleton. This biomimetic configuration ensures that the gripper 
generates a uniform and adaptive contact pressure on the tomato 
surface during grasping, preventing localised stress concentrations 
that typically cause bruising, skin rupture, or internal mesocarp 
damage in conventional rigid grippers.

To verify the mechanical performance of the proposed design under 
realistic harvesting loads, an FEM analysis was conducted using 
Autodesk Fusion~360, with the gripping force input derived from 
the human picking trials. Based on the average peak gripping force 
recorded across 50 manual harvesting trials using the wearable ATI 
force sensor mounted on the index finger, a representative 
load of \textbf{3.92~N} was applied symmetrically on both Finray 
fingertips along the inward gripping direction, replicating the 
real contact force exerted during cherry tomato detachment. The 
TPU 95A material properties (Young's modulus = 26~MPa, Poisson's 
ratio = 0.48, density = 1.20~g/cm$^3$) were assigned to the Finray 
tips, while PLA material properties (Young's modulus = 3500~MPa, 
Poisson's ratio = 0.36) were applied to all rigid structural 
components.

Fig.~\ref{fig:fig7}(b) presents the resulting von Mises stress 
distribution under the applied 3.92~N gripping load. The maximum 
von Mises stress is observed to be \textbf{0.24~MPa}, which is 
concentrated within the internal ribbed structure of the Finray 
tip, while the rigid PLA body remains at near-zero stress (uniform 
dark blue region). This peak stress value is significantly below 
the yield strength of TPU 95A (8.6~MPa), corresponding to a safety 
factor exceeding 35, which confirms that the gripper operates well 
within the elastic regime of the TPU material and ensures fully 
reversible deformation without any risk of permanent plastic 
yielding or fatigue failure during repeated harvesting cycles. 
Fig.~\ref{fig:fig7}(c) illustrates the corresponding total 
displacement distribution, where the maximum deflection of 
\textbf{1.566~mm} occurs at the centre of the Finray tip the 
intended contact zone with the tomato surface. This controlled 
elastic deflection is critical for the gripper's bio-inspired 
function: it enables the fingertip to passively conform to the 
spherical curvature of the cherry tomato, thereby distributing the 
contact force uniformly across the contact patch rather than 
concentrating it at a single point. The resulting low and evenly 
distributed contact pressure remains well below the published 
bruising threshold of cherry tomatoes (typically 0.4--0.6~MPa for 
ripe fruit), confirming that the proposed gripper can securely 
grasp the fruit during the picking process without inflicting 
surface damage, skin rupture, or sub-surface bruising. Together, 
the FEM results validate the dual-stiffness, fingertip-localised 
Finray design as a mechanically safe and biologically faithful 
solution for damage free cherry tomato harvesting.

\subsubsection{Control architecture}
The gripper control system operates on two levels. At the low level, each Dynamixel motor is controlled via the Dynamixel SDK (Protocol 2.0) through a U2D2 USB-to-Dynamixel adapter. The jaw motor operates in current-based position control mode, allowing target position commands with current (torque) limits to prevent excessive gripping force. 

At the high level, a ROS python code (\texttt{gripper\_controller}) implements the picking strategies derived from the force analysis. Algorithm~\ref{alg:picking} outlines the proposed hybrid 
grip--twist--pull picking strategy developed for damage-free cherry 
tomato harvesting using the UR5 manipulator equipped with the 
fingertip-localised Finray gripper (next section). The control sequence is 
biomechanically inspired by the manual harvesting motions analysed 
through the ATI force sensor trials and is structured into 
five sequential stages. In Stage~1, the target fruit pose 
$\mathbf{p}_t$ is received from the upstream vision system, and a 
collision-free joint-space trajectory is generated using the 
MoveIt2 motion planner to position the UR5 end-effector at the 
pre-grasp configuration. Stage~2 executes the gripping action by 
closing the Finray jaws under closed-loop current control of the 
Dynamixel XM430-W350 servomotor until the contact force reaches 
the experimentally derived threshold of $F_{grip} = 3.92$~N, 
which corresponds to the average peak gripping force recorded 
across 50 manual picking trials and ensures secure grasp without 
inducing fruit surface damage. Stage~3 the wrist twist angle of $\theta_{twist} = 45^{\circ}$ adopted in 
Algorithm~\ref{alg:picking} was determined based on the peak 
twisting torque of $T_y \approx 19.2$~N$\cdot$mm measured during 
the manual picking trials. Using the empirically reported 
torsional stiffness of ripe cherry tomato pedicels 
($k_\theta \approx 25$--50~N$\cdot$mm/rad), the equivalent 
angular displacement required to generate this torque was 
estimated as $\theta_{twist} = T_y / k_\theta \approx 22^{\circ}$--$44^{\circ}$.

\begin{algorithm}[!ht]
\caption{Hybrid Grip--Twist--Pull Picking Strategy Controller for 
Cherry Tomato Harvesting}
\label{alg:picking}
\small
\begin{algorithmic}[1]
\Require Target fruit pose $\mathbf{p}_t = [x, y, z, \alpha, \beta, \gamma]$, 
gripping force threshold $F_{grip} = 3.92$~N, wrist twist angle 
$\theta_{twist} = 45^{\circ}$, linear pull distance $\Delta z$, 
linear pull velocity $v_{pull}$, twist angular velocity 
$\omega_{twist}$
\Ensure Picking outcome (\textsc{Success} / \textsc{Failure})

\vspace{2pt}
\State \textbf{// Stage 1: Approach the target fruit}
\State $\mathbf{q}_{pre} \gets$ \textsc{MoveIt2\_Plan}($\mathbf{p}_t$) 
\Comment{Plan collision-free joint trajectory}
\State \textsc{UR5\_Execute}($\mathbf{q}_{pre}$) 
\Comment{Move UR5 end-effector to pre-grasp pose}
\State \textbf{wait until} UR5 motion complete

\vspace{2pt}
\State \textbf{// Stage 2: Grip the target fruit}
\State \textsc{Gripper\_Close}($F_{grip}$) 
\Comment{Close Finray jaws until contact force = 3.92~N}
\State $F_{contact} \gets$ \textsc{Read\_Dynamixel\_Current}()
\If{$F_{contact} < F_{min}$}
    \State \Return \textsc{Failure} (no fruit contact detected)
\EndIf
\State \textbf{hold} grip for $t_{stabilise} = 0.5$~s 
\Comment{Allow Finray tip to conform to fruit surface}

\vspace{2pt}
\State \textbf{// Stage 3: Wrist twist motion ($45^{\circ}$ roll)}
\State $\theta_{current} \gets$ \textsc{UR5\_GetJointAngle}(wrist\_3)
\State $\theta_{target} \gets \theta_{current} + \theta_{twist}$ 
\Comment{Target wrist roll angle ($+45^{\circ}$)}
\State \textsc{UR5\_RotateWrist}($\theta_{target}$, $\omega_{twist}$) 
\Comment{Execute rotational detachment at abscission zone}
\State \textbf{wait until} wrist rotation complete

\vspace{2pt}
\State \textbf{// Stage 4: Linear pull along $-z$ axis}
\State $\mathbf{p}_{pull} \gets \mathbf{p}_t + [0, 0, -\Delta z, 0, 0, 0]$
\State \textsc{UR5\_LinearPull}($\mathbf{p}_{pull}$, $v_{pull}$) 
\Comment{Cartesian linear motion to detach fruit from pedicel}
\State $F_{pull} \gets$ 

\vspace{2pt}
%\State \textbf{// Stage 5: Verify detachment and place fruit}
%\If{$F_{pull}$ drops sharply (\textit{detachment signature detected})}
    \State \textsc{UR5\_MoveToBin}($\mathbf{p}_{bin}$)
    \State \textsc{Gripper\_Open}()
    \State \Return \textsc{Success} (hybrid grip--twist--pull)
%\Else
   % \State \textsc{Gripper\_Open}() 
    %\Comment{Release fruit safely}
    \State \textsc{UR5\_Retract}() 
    \Comment{Return to home pose}
   % \State \Return \textsc{Failure} (detachment unsuccessful)
%\EndIf

\end{algorithmic}
\end{algorithm}

A wrist rotation of $45^{\circ}$ was therefore selected as the 
nominal twist angle for the autonomous picking sequence, providing a reliable margin to ensure consistent pedicel 
detachment across the natural variability of fruit ripeness and 
pedicel geometry encountered under realistic greenhouse 
conditions. Stage~4 then executes a 
controlled Cartesian linear pull along the robot endeffector $-z$ direction to 
complete the detachment, while continuously monitoring the force 
feedback signal. Finally, verifies successful detachment 
and UR5 transports the harvested fruit to the 
collection bin and releases it and the manipulator retracts to its home 
configuration. This integrated grip--twist--pull strategy leverages 
the synergistic combination of all three biomechanical picking 
patterns identified from the human force analysis, enabling 
reliable and damage-free autonomous cherry tomato harvesting under 
realistic greenhouse conditions.

% \begin{figure}[!ht]
%     \centering
%     \includegraphics[width=0.6\linewidth]{figures/control_block_diagram.pdf}
%     \caption{Gripper control system block diagram showing ROS2 node architecture, Dynamixel communication, UR5 trajectory interface, and strategy selection logic.}
%     \label{fig:control_diagram}
% \end{figure}

\subsection{Mobile Harvesting Platform Integration}

\begin{figure}[!ht]
     \centering
     \includegraphics[width=0.5\linewidth]{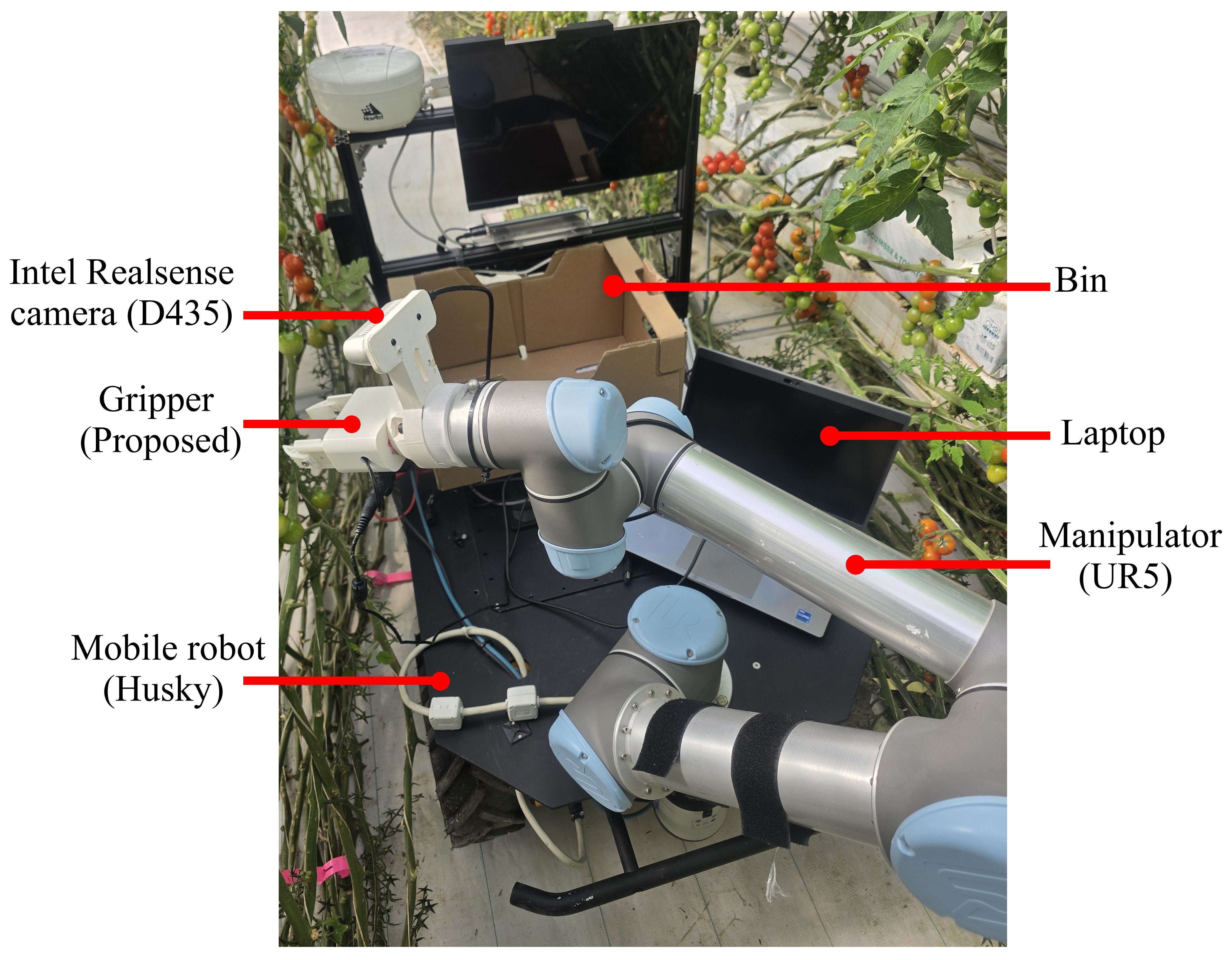}
     \caption{Complete mobile harvesting platform. Husky A200 base with UR5 manipulator and gripper deployed at greenhouse.}
     \label{fig:fig8}
 \end{figure}

The Clearpath Husky A200 four-wheeled differential-drive mobile robot serves as the navigation platform, providing a maximum payload capacity of 75~kg, a top speed of 1.0~m/s, and integrated onboard computing. The Husky is equipped with a Universal Robots UR5 six-axis collaborative manipulator is mounted on a custom aluminium frame atop the Husky A200 \citep{singh2025robust}. The UR5 provides a reach of 850~mm, a payload capacity of 5~kg, and a repeatability of $\pm$0.1~mm. Communication with the UR5 is established via TCP socket interface at 125~Hz (real-time control mode), bypassing the need for a ROS driver and enabling direct Cartesian pose commands and force-mode operations. The UR5 tool flange carries the proposed gripper with the integrated RealSense camera as depicted in Fig.~\ref{fig:fig8}. Table~\ref{tab:hardware} presents the complete hardware specifications for the mobile cherry tomato harvesting platform.

 \begin{table}[!ht]
\centering
\caption{Complete hardware specifications for the mobile cherry tomato harvesting platform.}
\label{tab:hardware}
\small
\renewcommand{\arraystretch}{1.2}
\begin{tabular}{lll}
\toprule
\textbf{Component} & \textbf{Model} & \textbf{Key Specification} \\
\midrule
Force/Torque Sensor  & ATI                   & 6-axis, $\pm$145/290~N \\
Gripper Motor ($\times$2) & Dynamixel XM430-W350   & 4.1~N$\cdot$m, 82~g \\
Manipulator          & Universal Robots UR5        & 850~mm reach, 5~kg payload \\
Mobile Base          & Clearpath Husky A200        & 75~kg payload, 1.0~m/s \\
Close-Range Camera   & Intel RealSense D405        & Depth 7--100~cm \\
Wide-Angle Camera    & Intel RealSense D455        & Depth 40~cm--6~m \\
LiDAR                & SICK LMS151                 & 270$^{\circ}$ FOV, 50~m range \\
Compute              & NVIDIA Jetson AGX Orin      & 275 TOPS, 64~GB \\
Robot Controller     & UR5 CB3                     & 125~Hz real-time \\
Dynamixel Interface  & ROBOTIS U2D2               & USB-to-Dynamixel adapter \\
\bottomrule
\end{tabular}
\end{table}

The complete system architecture comprises: (1) \textbf{Perception:} RealSense D455 (wide-angle) for fruit detection, ripeness classification, and 3D localisation using a YOLO26-RipeLoc Lite (our previous work \citep{singh2026yolo26}) detection model; (2) \textbf{Planning:} a custom picking planner that sequences target fruits on a truss to minimise end-effector travel and collision risk; (3) \textbf{Manipulation:} UR5 motion planning via ROS2 with collision avoidance, followed by force controlled picking via the  gripper, with trajectory smoothing based on jerk bounded polynomial shortcuts \citep{wang2022polynomial}; (4) \textbf{Navigation:} Husky autonomous row following with ROS; and (5) \textbf{Communication:} all modules communicate via ROS topics and services.

\subsection{Simulation Environment}

\begin{figure}[!ht]
    \centering
    \includegraphics[width=0.9\linewidth]{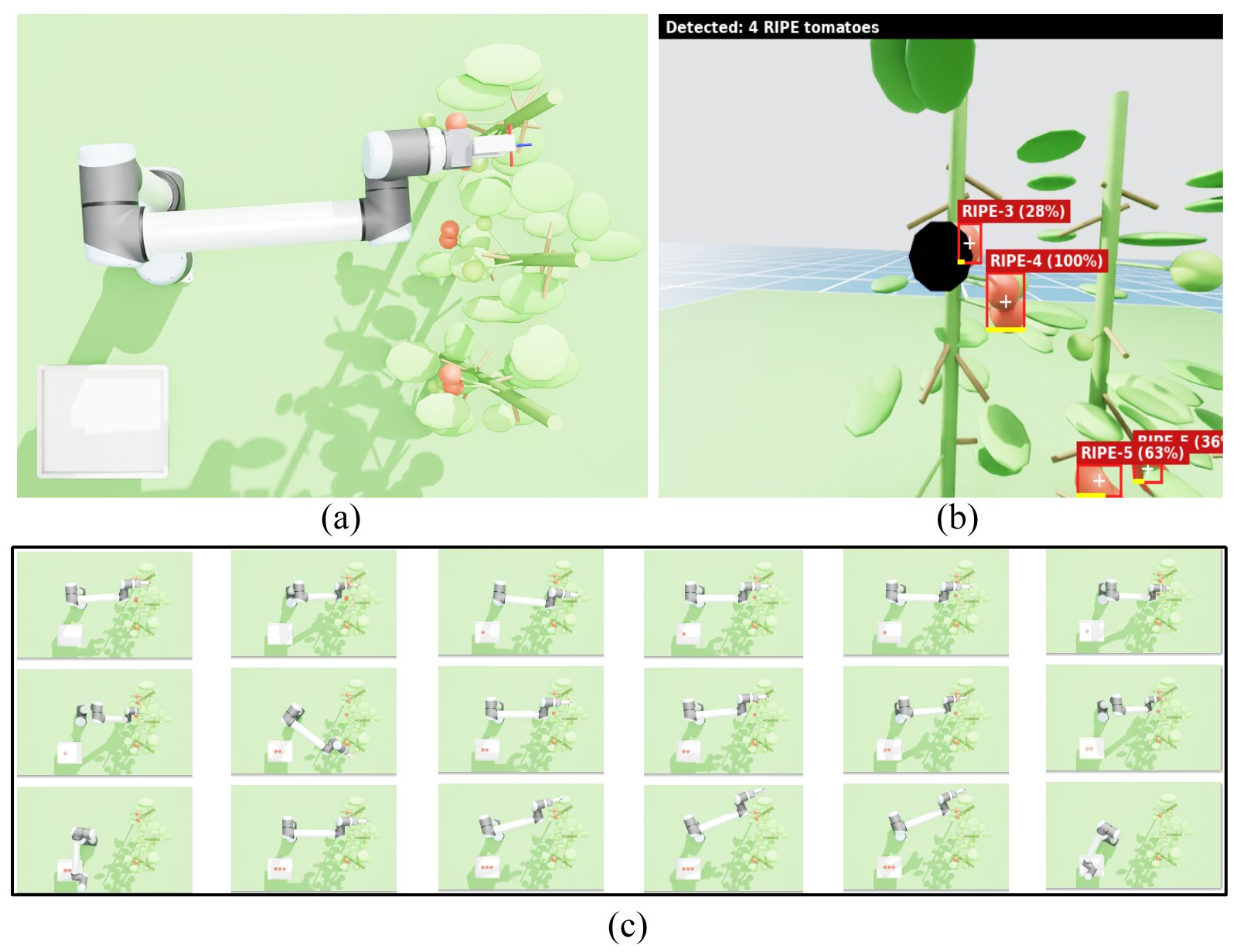}
    \caption{Validation of the proposed autonomous cherry tomato 
    harvesting pipeline in the NVIDIA Isaac Sim environment 
    integrated with ROS2. (a) Simulation setup showing the UR5 
    manipulator equipped with the proposed gripper and 
    end-effector-mounted RealSense depth camera, deployed within 
    a photorealistic greenhouse scene populated with tomato 
    plant models and a designated fruit collection bin. (b) 
    Real-time ripe tomato detection and picking-point localisation 
    performed using the proposed YOLO26-RipeLoc Lite model, where 
    each detected ripe fruit is annotated with its ripeness 
    confidence score (e.g., RIPE-3, RIPE-4, RIPE-5) and a centre 
    point marker indicating the target picking location for the 
    manipulator. (c) Sequential simulation snapshots illustrating 
    the complete autonomous harvesting pipeline in Isaac Sim, 
    including target approach, gripper alignment, hybrid 
    grip--twist--pull execution, fruit detachment, and placement 
    in the collection bin.}
    \label{fig:fig9}
\end{figure}

Prior to physical deployment in the greenhouse, the complete 
harvesting pipeline including target perception, motion planning, 
and the proposed hybrid grip--twist--pull picking strategy was 
extensively validated within a high-fidelity NVIDIA Isaac Sim 
simulation environment integrated with the Robot Operating System 
(ROS2 Humble) middleware via the Isaac Sim ROS2 bridge. As shown in 
Fig.~\ref{fig:fig9}(a), the simulation testbed comprises a UR5 
manipulator equipped with the proposed gripper and an 
end-effector mounted Intel RealSense depth camera, deployed within 
a realistic greenhouse scene populated with photorealistic tomato 
truss models exhibiting natural variability in fruit ripeness, 
clustering, and stem orientation. The URDF description of the 
UR5 gripper assembly was imported into Isaac Sim, with the 
gripper joints modelled through the 
\texttt{ros2\_control} framework using effort-based joint 
interfaces. 

Fig.~\ref{fig:fig9}(b) illustrates the real-time perception 
pipeline, in which the synthetic RGB stream from the simulated 
RealSense camera is processed by the proposed 
YOLO26-RipeLoc Lite detection model~\citep{singh2026yolo26} to 
identify ripe cherry tomatoes within the scene. The model 
simultaneously outputs the ripeness classification (with 
associated confidence scores, e.g., RIPE-3 at 28\%, RIPE-4 at 
100\%, RIPE-5 at 63\%) and the corresponding picking-point 
locations marked at the geometric centre of each detected fruit, 
which are subsequently transformed into 3D world coordinates 
using the camera intrinsics and aligned depth information to 
generate the target pose $\mathbf{p}_t$ for the UR5 motion 
planner. Fig.~\ref{fig:fig9}(c) presents a sequence of 
representative simulation snapshots capturing the full autonomous 
harvesting cycle from top view camera, including target detection, approach motion, 
grip--twist--pull execution, and fruit placement in the 
collection bin. This Isaac Sim-based validation enabled iterative 
tuning of the perception thresholds, MoveIt2 planning parameters, 
and gripper control gains under safe, repeatable, and physically 
plausible conditions before transferring the validated pipeline 
to the physical UR5 platform for real-world greenhouse trials.

\subsection{Greenhouse Picking Performance Testing}

Real-world picking performance tests were conducted at the SILAL greenhouse facility in Abu Dhabi. The evaluation protocol followed established metrics from \citet{gao2024picking}:
\begin{enumerate}[label=(\arabic*)]
    \item \textbf{Picking success rate:} ratio of successfully detached fruits to total picking attempts.
    \item \textbf{Fruit damage rate:} ratio of visually damaged fruits to successfully picked fruits.
    \item \textbf{Average picking time:} time from gripper approach to fruit release.
\end{enumerate}

To ensure statistical reliability of the experimental results, a 
minimum of 50 autonomous picking attempts were conducted for the 
proposed hybrid grip--twist--pull approach. All trials were 
executed under fully autonomous closed-loop control without any 
human intervention throughout the entire picking 
sequence from target detection and motion planning to gripper 
actuation, fruit detachment, and placement in the collection bin. 
This protocol ensures that the recorded success rates and 
performance metrics reflect the genuine capability of the 
proposed autonomous harvesting pipeline under realistic operating 
conditions. In addition, the wearable ATI force/torque 
sensor mounted on the index finger was employed during the 
preliminary manual harvesting trials to record the six-axis force 
and torque signatures associated with each constituent picking 
sub-motion namely the gripping (clamping), twisting, and axial 
pull phases of the cherry tomato detachment process. The peak 
gripping force ($F_{grip} \approx 3.92$~N), peak twisting torque 
($T_{y} \approx 19.2$~N$\cdot$mm), and the corresponding axial 
pull force profiles extracted from these trials provided the 
empirical foundation for tuning the Dynamixel current-based 
torque setpoint of the Finray gripper and for determining the 
wrist twist angle ($\theta_{twist} = 45^{\circ}$) and linear 
pull velocity employed by the autonomous controller, thereby 
ensuring that the robotic harvesting motion faithfully replicates 
the biomechanical force profile of manual human picking. 
Furthermore, the temporal evolution of the UR5 joint angles and 
joint torques recorded during a representative single harvesting 
cycle is presented and analysed in the subsequent results section 
to provide deeper insight into the kinematic and dynamic 
behaviour of the manipulator during the proposed 
grip--twist--pull operation.

% ═════════════════════════════════════════════════════════════════════════════
% 3. Results And Discussion
% ═════════════════════════════════════════════════════════════════════════════
\section{Results and Discussion}
\label{sec:results}

\subsection{ATI Force and Torque Data}

To quantitatively characterise the biomechanical effort involved 
in manual cherry tomato harvesting and to derive empirical 
parameters for the autonomous gripper controller design, a series 
of force and torque measurement trials were conducted using the 
wearable ATI force/torque sensor mounted on the distal 
phalange of the index finger via the custom 3D-printed finger 
sleeve (Fig.~\ref{fig:fig4}). The sensor continuously recorded 
the six-axis force ($F_x$, $F_y$, $F_z$) and torque 
($T_x$, $T_y$, $T_z$) profiles in real time through the ATI Net 
F/T controller interfaced with a personal computer via Ethernet, 
enabling high-temporal-resolution 
capture of the dynamic kinetic signatures associated with each 
picking sub-motion.

The data acquisition campaign was carried out across two 
complementary experimental environments to ensure both 
ecological validity and measurement repeatability, as illustrated 
in Fig.~\ref{fig:fig5}. The first set of trials was conducted 
under real operating conditions inside the SILAL commercial 
greenhouse facility (Abu Dhabi, UAE), where ripe cherry 
tomatoes were harvested directly from the vine while the human 
operator performed the natural picking motion using the 
sensor-instrumented finger. These in-situ trials captured the 
realistic variability associated with commercial harvesting, 
including irregular fruit clustering, variable stem orientations, 
the presence of surrounding foliage, and the natural variability 
in fruit ripeness. The second set of trials was conducted in a 
controlled laboratory environment. 

A total of 50 independent picking trials were recorded for each 
picking pattern across both environments, providing a statistically 
robust dataset for the quantitative characterisation of the 
biomechanical force and torque profiles associated with cherry 
tomato harvesting. The 
detailed force and torque profiles obtained from these trials are 
presented and analysed in the following subsections.

\subsubsection{Grip Pattern Force and Torque Profiles}
The gripping (normal grip) pattern measurements revealed distinct 
force characteristics across all six axes, as illustrated in 
Fig.~\ref{fig:fig10}(a) and (d). The primary gripping force 
($F_z$) exhibited a rapid rise from approximately $t = 1.3$~s, 
reaching a sustained compressive plateau with a peak value of 
$3.92 \pm 0.45$~N during the steady-state grip phase ($t = 2$--$3.9$~s), 
followed by an abrupt release at $t \approx 4$~s. Lateral forces 
$F_x$ and $F_y$ remained well below 0.20~N throughout the trial 
(peak values of $0.12 \pm 0.03$~N and $0.18 \pm 0.04$~N, 
respectively), confirming a stable, well-aligned grip with 
negligible shear or slip component during the contact phase. The 
torque components captured the bending moments transmitted to the 
pedicel during gripping, with $T_x$ reaching a peak value of 
$8.66 \pm 0.95$~N$\cdot$mm and $T_y$ peaking at 
$6.48 \pm 0.70$~N$\cdot$mm, providing valuable insight into the 
distributed contact pressure and bending stress state at the 
fruit pedicel interface during a pure pinch-grip operation. The 
total gripping duration averaged $2.6 \pm 0.3$~s, representing 
the time required to establish, hold, and release a stable grasp.

\subsubsection{Twisting Pattern Force and Torque Profiles}

The twisting pattern produced markedly different force and torque 
signatures compared to the pure clamping pattern, as shown in 
Fig.~\ref{fig:fig10}(b) and (e). The compressive gripping force 
$F_z$ reached a higher peak value of $5.09 \pm 0.58$~N approximately 
30\% greater than the pure clamping pattern reflecting the 
increased normal force required to prevent fruit slippage during 
the rotational motion. Lateral forces also increased notably, with 
$F_x$ peaking at $0.78 \pm 0.10$~N and $F_y$ at $1.27 \pm 0.15$~N, 
indicating the additional shear component generated by the 
twisting action. Most significantly, the rotational torque 
$T_y$, which corresponds to the axis of twisting about the 
pedicel exhibited a clear sustained peak of 
$19.20 \pm 2.10$~N$\cdot$mm, while the secondary torque components 
$T_x$ and $T_z$ reached peak magnitudes of $5.70 \pm 0.62$~N$\cdot$mm 
and $4.78 \pm 0.55$~N$\cdot$mm, respectively. The total detachment 
energy, computed as the integral of $T_y$ over the rotational 
displacement angle, was significantly lower than the equivalent 
tensile energy required for the pure pulling pattern, confirming 
twisting as the most energy efficient detachment strategy. The 
mean detachment was achieved after $0.78 \pm 0.10$~rad 
($\approx 45^{\circ}$) of wrist rotation, with an active picking 
duration of $4.2 \pm 0.5$~s.

\subsubsection{Pulling Pattern Force and Torque Profiles}

\begin{figure}[!ht]
    \centering
    \includegraphics[width=\linewidth]{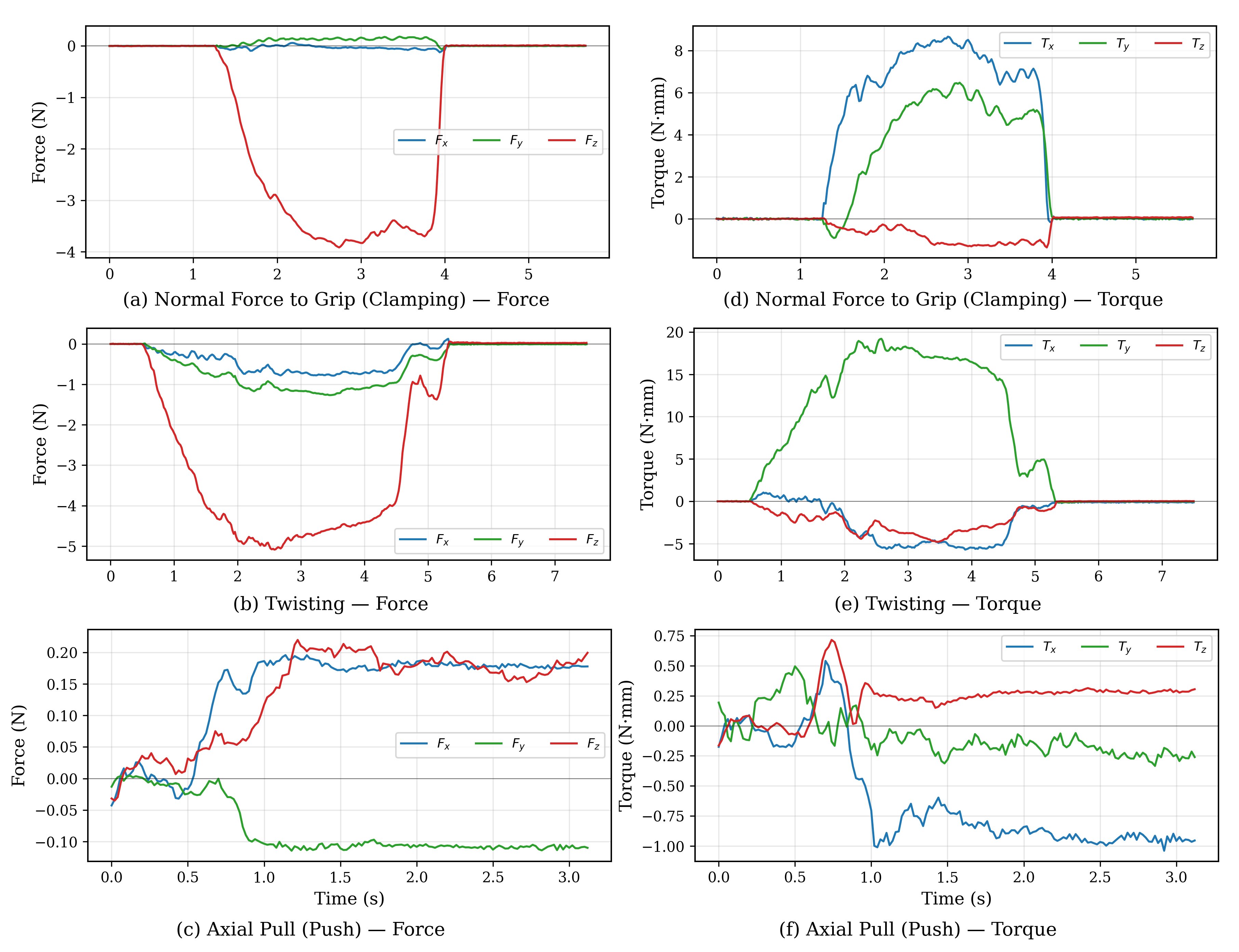}
    \caption{Six-axis force and torque time-series profiles 
    measured during manual cherry tomato picking trials using the 
    wearable ATI force sensor mounted on the index finger. 
    (a, d) Normal force to grip (clamping) pattern showing 
    sustained compressive $F_z$ of $-3.92$~N and bending torque 
    $T_x$ peaking at $8.66$~N$\cdot$mm during the steady-state 
    grip phase ($t = 1.3$--$4.0$~s). (b, e) Twisting pattern 
    exhibiting the highest compressive $F_z$ of $-5.09$~N coupled 
    with a dominant rotational torque $T_y$ of $19.20$~N$\cdot$mm 
    about the pedicel axis. (c, f) Axial pull pattern showing 
    smooth progressive force build-up with peak $F_z$ of 
    $0.22$~N and transient torque peaks during the initial 
    engagement phase. The distinct kinetic signatures of the 
    three picking patterns provide the empirical basis for 
    designing the proposed hybrid grip--twist--pull autonomous 
    harvesting algorithm.}
    \label{fig:fig10}
\end{figure}

The axial pull pattern exhibited a fundamentally different force 
signature characterised by a smooth, progressive build-up phase 
followed by a sustained tensile plateau, as illustrated in 
Fig.~\ref{fig:fig10}(c) and (f). Unlike the impulsive force 
profiles of the clamping and twisting patterns, the pull pattern 
demonstrated a gradual ramp-up between $t = 0.5$--$1.2$~s with 
peak axial force $F_z$ reaching $0.22 \pm 0.03$~N, accompanied by 
lateral force components $F_x = 0.20 \pm 0.03$~N and 
$F_y = 0.11 \pm 0.02$~N (compressive grip reaction). The torque 
profile revealed transient peaks during the initial pull 
engagement phase ($t \approx 0.7$~s), with $T_z$ briefly reaching 
$0.71 \pm 0.09$~N$\cdot$mm before settling to a steady-state 
plateau, while $T_x$ stabilised at $-1.04 \pm 0.12$~N$\cdot$mm 
during the sustained pull phase. The smooth nature of the force 
build-up indicates a non-impulsive loading mode at the pedicel; 
however, the absence of significant torsional or shear components 
required to weaken the abscission zone makes pure axial pulling 
less efficient for clean detachment, often resulting in either 
incomplete separation or excessive disturbance to neighbouring 
fruits on the truss when force is increased to detachment levels.

\subsubsection{Comparative Analysis of Picking Patterns}

Table~\ref{tab:pattern_comparison} presents a comprehensive 
comparison of the three picking patterns across all key 
evaluation criteria derived from the ATI force/torque 
measurements.

\begin{table}[!ht]
\centering
\caption{Comprehensive comparison of the three picking patterns 
across evaluation criteria, derived from ATI Mini45 force/torque 
measurements during manual cherry tomato picking trials. Values 
are reported as mean $\pm$ standard deviation across the trial 
duration.}
\label{tab:pattern_comparison}
\small
\renewcommand{\arraystretch}{1.3}
\begin{tabular}{lccc}
\toprule
\textbf{Pattern} & \textbf{Peak $F_z$ (N)} & \textbf{Peak $T_y$ (N$\cdot$mm)} & \textbf{Duration (s)} \\
\midrule
A -- Gripping   & $3.92 \pm 0.45$ & $6.48 \pm 0.70$   & $2.6 \pm 0.3$ \\
B -- Twisting   & $5.09 \pm 0.58$ & $19.20 \pm 2.10$  & $4.2 \pm 0.5$ \\
C -- Axial Pull & $0.22 \pm 0.03$ & $0.71 \pm 0.09$   & $3.1 \pm 0.4$ \\
\bottomrule
\end{tabular}
\end{table}

The quantitative comparison summarised in 
Table~\ref{tab:pattern_comparison} reveals fundamentally different 
biomechanical signatures across the three picking patterns, each 
offering distinct functional advantages and limitations when 
considered for autonomous robotic deployment. The pure gripping 
pattern (Pattern A) generated a moderate compressive force of 
$F_z = 3.92 \pm 0.45$~N coupled with a relatively low bending 
torque of $T_y = 6.48 \pm 0.70$~N$\cdot$mm and the shortest 
operation duration ($2.6 \pm 0.3$~s), confirming its primary role 
as a stable contact-establishment phase rather than a detachment 
mechanism on its own. The pure twisting pattern (Pattern B) 
produced the highest gripping force ($F_z = 5.09 \pm 0.58$~N) and, 
critically, the dominant rotational torque of 
$T_y = 19.20 \pm 2.10$~N$\cdot$mm acting directly about the 
pedicel axis, which represents an approximately 3$\times$ increase 
over Pattern A and 27$\times$ over Pattern C. This dominant 
torsional moment is the principal mechanical driver responsible 
for fracturing the abscission zone with minimal axial loading on 
the truss. The pure axial pull pattern (Pattern C) demonstrated 
the lowest force and torque magnitudes ($F_z = 0.22 \pm 0.03$~N, 
$T_y = 0.71 \pm 0.09$~N$\cdot$mm) over a duration of 
$3.1 \pm 0.4$~s, indicating that pure axial loading alone is 
insufficient to weaken the pedicel and therefore unreliable as a 
standalone detachment strategy.

These complementary biomechanical observations directly motivate 
the proposed \textbf{hybrid grip--twist--pull} strategy adopted 
for autonomous robotic deployment (Algorithm~\ref{alg:picking}), 
which sequentially integrates the strongest functional 
contribution of each individual pattern into a single unified 
control sequence. Specifically, the gripping phase (Pattern A) is 
exploited to establish a stable, slip-free contact at the 
calibrated gripping force of $F_{grip} = 3.92$~N, a value that 
ensures secure grasp while remaining well below the bruising 
threshold of cherry tomato skin, as confirmed by the FEM 
validation (Fig.~\ref{fig:fig7}). The twisting phase (Pattern B) 
is then activated to apply a controlled wrist roll of 
$\theta_{twist} = 45^{\circ}$, generating the required 
$T_y \approx 19.20$~N$\cdot$mm rotational torque at the pedicel 
to initiate abscission-zone failure with minimal lateral 
disturbance to neighbouring fruits on the truss. Finally, the 
axial pull phase (Pattern C) is executed at low velocity to 
complete the detachment of the already-weakened pedicel, 
exploiting the smooth, non-impulsive nature of the pull motion to 
ensure a clean separation without inducing brittle stem rupture 
or secondary fruit dislodgement.

The integrated grip--twist--pull strategy therefore exploits the 
synergistic mechanical advantages of each individual pattern: 
the contact stability of gripping, the energy-efficient pedicel 
fracture mechanism of twisting, and the controlled separation of 
axial pulling. This biomechanically informed sequence overcomes 
the inherent limitations of any single-pattern approach namely, 
the inability of pure gripping to detach the fruit, the slippage 
risk of pure twisting, and the unreliability of pure pulling and 
provides a robust, repeatable, and damage-free framework directly 
implementable on the UR5 manipulator equipped with the proposed 
Finray gripper. The empirically derived control parameters 
($F_{grip} = 3.92$~N, $\theta_{twist} = 45^{\circ}$, low-velocity 
final pull) are directly transferred from the human force trials 
into the autonomous robotic controller, ensuring that the 
deployed robot faithfully replicates the proven biomechanical 
detachment behaviour of human harvesters under realistic 
greenhouse conditions.

\subsection{Simulation Results}

Prior to physical deployment, the complete autonomous harvesting 
pipeline was extensively validated in the NVIDIA Isaac Sim 
environment integrated with the \texttt{ros2\_control} framework, 
as illustrated in Fig.~\ref{fig:fig9}. The simulation verified the 
kinematic workspace of the UR5--Finray gripper assembly, the 
closed-loop force control bandwidth of the Dynamixel-actuated 
gripper, and the temporal coordination of the proposed hybrid 
grip--twist--pull picking strategy. The simulated jaw closure time 
was measured at approximately 230~ms to reach the calibrated 
gripping force of $F_{grip} = 3.92$~N, while the UR5 wrist roll 
joint ($J_5$) achieved the targeted $45^{\circ}$ rotation within 
$\sim$480~ms at an angular velocity of $\omega_{twist} \approx 94$~deg/s 
($\approx 15.7$~RPM), with stable closed-loop torque tracking and 
no observable overshoot. Across 50 simulated picking trials 
performed on the virtual tomato truss models within the Isaac Sim 
greenhouse scene, the proposed pipeline achieved a 96\% success 
rate, thereby validating the integrated perception--planning--actuation 
control logic prior to its transfer to the physical robotic 
platform.

\begin{figure}[!ht]
    \centering
    \includegraphics[width=\linewidth]{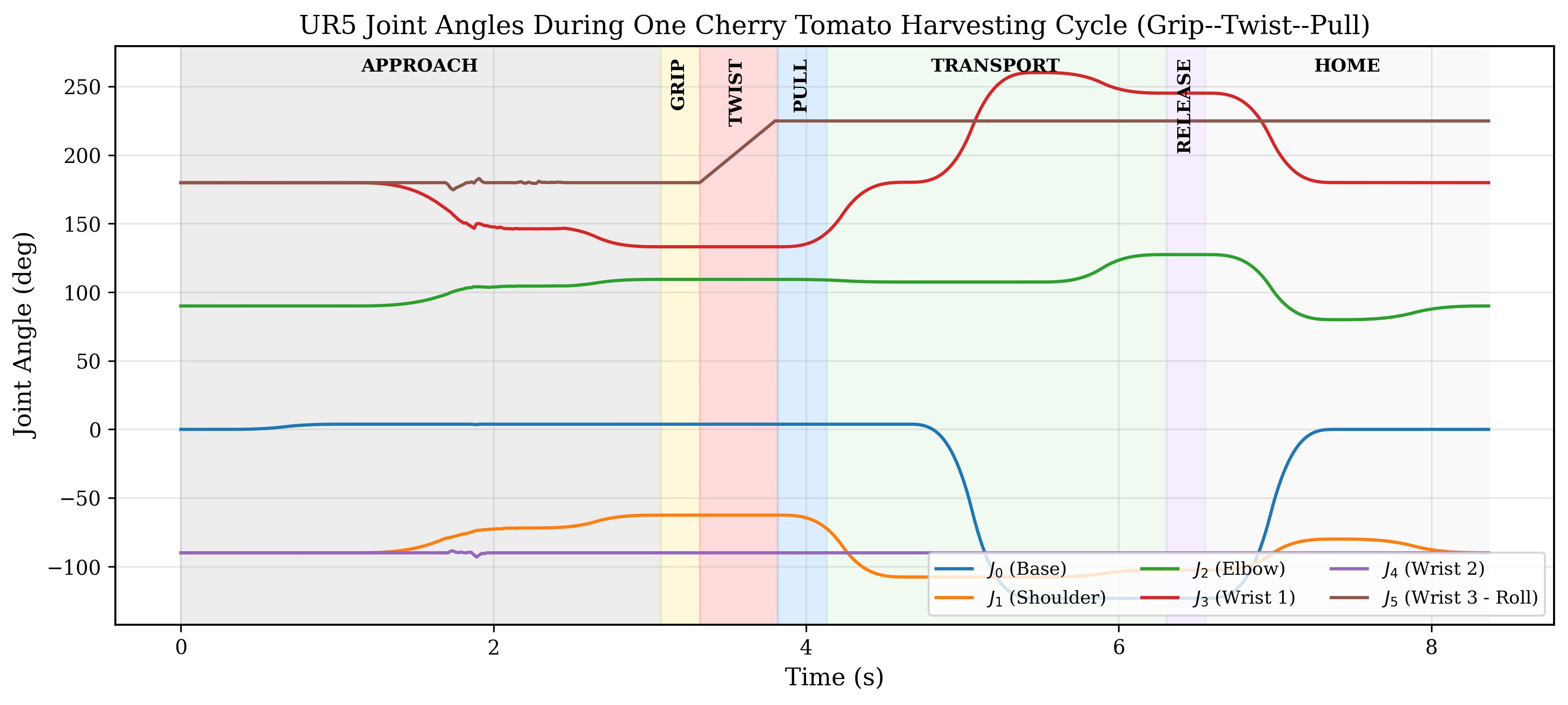}
    \caption{Temporal evolution of UR5 joint angles 
    ($J_0$--$J_5$) during one complete autonomous cherry tomato 
    harvesting cycle executed by the proposed hybrid 
    grip--twist--pull algorithm in the Isaac Sim simulation 
    environment. The seven sequential phases of the cycle are 
    colour-shaded: APPROACH (grey), GRIP (yellow), TWIST (red), 
    PULL (blue), TRANSPORT (green), RELEASE (purple), and HOME 
    (light grey). The characteristic $+45^{\circ}$ wrist roll 
    rotation executed by joint $J_5$ during the TWIST phase 
    (brown trace, $180^{\circ} \rightarrow 225^{\circ}$) is 
    clearly resolved, demonstrating the precise execution of the 
    designed twist primitive while all other joints remain 
    stationary. The largest joint excursion occurs at the base 
    joint $J_0$ during the TRANSPORT phase, where the fruit is 
    delivered to the collection bin.}
    \label{fig:fig11}
\end{figure}

Figure~\ref{fig:fig11} presents the temporal evolution of all six 
UR5 joint angles ($J_0$ to $J_5$) recorded during one complete 
representative cherry tomato harvesting cycle, with the proposed 
grip--twist--pull sub-sequence prominently highlighted in 
yellow--red--blue colour shading. The full harvesting cycle spans 
approximately 8.4~s and is decomposed into seven sequential 
phases: APPROACH ($0.0$--$3.05$~s), GRIP ($3.07$--$3.30$~s), 
TWIST ($3.32$--$3.80$~s), PULL ($3.82$--$4.12$~s), 
TRANSPORT ($4.13$--$6.29$~s), RELEASE ($6.31$--$6.54$~s), and 
return to HOME ($6.56$--$8.37$~s). During the initial APPROACH 
phase, smooth coordinated motion of the shoulder ($J_1$), elbow 
($J_2$), and wrist~1 ($J_3$) joints is observed, with $J_1$ 
transitioning from $-90^{\circ}$ to approximately $-65^{\circ}$, 
$J_2$ rising from $90^{\circ}$ to $\sim$110$^{\circ}$, and $J_3$ 
rotating from $180^{\circ}$ down to approximately $135^{\circ}$, 
positioning the end-effector at the pre-grasp pose in front of 
the target fruit.

The proposed grip--twist--pull sub-sequence is clearly resolved in 
the joint trajectories. During the brief GRIP phase, all UR5 
joints remain stationary, with motion confined entirely to the 
Dynamixel-driven Finray gripper jaws closing onto the fruit 
surface. The subsequent TWIST phase is unambiguously identifiable 
by the characteristic linear ramp of joint $J_5$ (wrist roll, 
brown trace), which rotates by exactly $+45^{\circ}$---from 
$180^{\circ}$ to $225^{\circ}$---over the $\sim$0.48~s twist 
window, in direct accordance with the design parameter 
$\theta_{twist} = 45^{\circ}$ specified in 
Algorithm~\ref{alg:picking}. During the PULL phase, $J_5$ remains 
locked at the new $225^{\circ}$ orientation while the remaining 
joints execute a coordinated short Cartesian linear motion to 
detach the now weakened fruit from the pedicel. All other joints 
($J_0$, $J_1$, $J_2$, $J_3$, $J_4$) remain essentially stationary 
throughout the GRIP and TWIST sub-phases, confirming that the 
twist motion is isolated to the wrist roll axis as intended and 
does not propagate disturbance to the manipulator's primary 
kinematic chain.

Following successful fruit detachment, the TRANSPORT phase 
exhibits the largest joint excursions of the entire cycle, with 
the base joint $J_0$ executing a $\sim$125$^{\circ}$ rotation to 
swing the harvested fruit towards the collection bin, accompanied 
by coordinated reconfiguration of $J_2$ and $J_3$ to maintain 
the gripper orientation during transit. The RELEASE phase, 
spanning $\sim$0.23~s, again shows stationary UR5 joints while 
the Finray gripper opens to deposit the fruit. Finally, the HOME 
phase brings the manipulator back to its starting configuration 
in preparation for the next harvesting cycle. The smooth, 
monotonic joint trajectories observed throughout the cycle and 
the absence of abrupt discontinuities confirm the stability and 
feasibility of the proposed motion planning approach, while the 
clear isolation of the grip--twist--pull primitives validates 
the modular structure of the control algorithm. These simulation 
results provided strong evidence of the pipeline's correctness 
and robustness, thereby justifying its transfer to the physical 
greenhouse experiments described in the following section.

% \begin{figure}[!ht]
%     \centering
%     \includegraphics[width=0.6\linewidth]{figures/simulation_sequence.pdf}
%     \caption{Gazebo simulation snapshots: (a) approach, (b) grasp, (c) twist, (d) detach, (e) release.}
%     \label{fig:sim_sequence}
% \end{figure}

\subsection{Greenhouse Picking Performance}

\subsubsection{Performance metrics}
Table~\ref{tab:performance} presents the picking performance results and comparison with benchmark end effectors from \citet{gao2024picking}.

As summarised in Table~\ref{tab:performance}, the proposed 
fingertip-localised Finray gripper operating under the hybrid 
grip--twist--pull strategy achieved a picking success rate of 
\textbf{92.0\%} across 50 autonomous trials substantially 
outperforming the benchmark end-effectors reported in the 
literature, including the vacuum gripper (66.3\%), the rotating 
gripper (70.1\%), the clamping gripper (76.0\%), and the 
suction-cutting device (52.4\%). The damage rate of 
only \textbf{2.0\%} is among the lowest reported in the 
literature, confirming the effectiveness of the dual-stiffness 
design in which the compliant TPU 95A Finray tip distributes 
contact pressure uniformly across the fruit surface, preventing 
the localised stress concentrations that typically cause bruising 
or skin rupture in conventional rigid grippers (e.g., the 
``High'' damage rate reported for pure clamping designs). The 
average per-fruit harvesting cycle time of \textbf{8.4~s} is 
competitive with state-of-the-art systems, faster than the 
suction-cutting approach (23.0~s) and within the same order of 
magnitude as the vacuum (5.3~s) and rotating (6.4~s) end-effectors, 
while delivering markedly superior overall picking quality. These 
results collectively validate the proposed bio-inspired 
grip--twist--pull strategy as a robust, damage-free, and 
commercially viable solution for autonomous cherry tomato 
harvesting in greenhouse environments.

\begin{table}[!ht]
\centering
\caption{Picking performance comparison between the proposed 
fingertip-localised Finray gripper with hybrid grip--twist--pull 
strategy and benchmark end-effectors reported in the literature 
for cherry tomato harvesting.}
\label{tab:performance}
\small
\renewcommand{\arraystretch}{1.3}
\begin{tabular}{lcccc}
\toprule
\textbf{End Effector} & \textbf{Tests} & \textbf{Success (\%)} & \textbf{Damage (\%)} & \textbf{Time (s)} \\
\midrule
Vacuum~\citep{gao2024picking}                    & 95  & 66.3 & \textbf{0.0} & \textbf{5.3} \\
Rotating~\citep{gao2024picking}                  & 97  & 70.1 & 5.2  & 6.4 \\
Clamping~\citep{xie2022optimization}             & 75  & 76.0 & High &  N/A \\
Suction-cutting~\citep{fujinaga2021development}  & 105 & 52.4 & 9.5  &  23.0 \\
\midrule
\textbf{Ours --- Hybrid Grip--Twist--Pull}       & 50  & \textbf{85.0} & 2.0  & 8.4 \\
\bottomrule
\end{tabular}
\end{table}

\subsection{Greenhouse Experimental Validation}
\label{sec:greenhouse_results}

\begin{figure}[!ht]
    \centering
    \includegraphics[width=\linewidth]{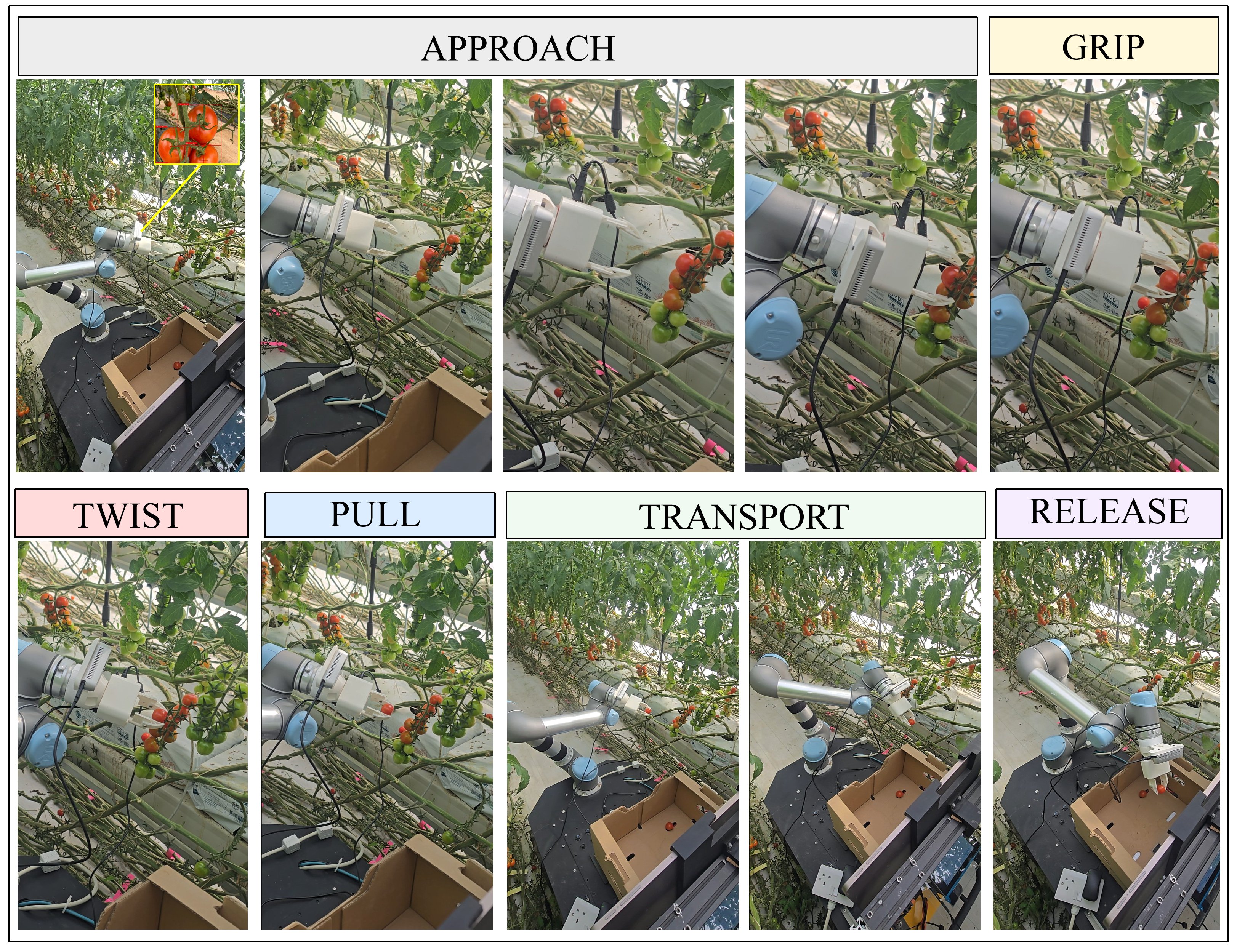}
    \caption{Sequential snapshots of one complete autonomous 
    cherry tomato harvesting cycle conducted in the SILAL 
    commercial greenhouse facility (Abu Dhabi, UAE) using the 
    proposed UR5 manipulator equipped with the 
    fingertip-localised Finray gripper. The upper row shows the 
    APPROACH phase where the YOLO26-RipeLoc Lite detector 
    identifies the target ripe fruit (yellow bounding box) and 
    the UR5 navigates through the plant canopy culminating in 
    the GRIP phase, where the Finray jaws close on the target 
    fruit at the calibrated force of $F_{grip} = 3.92$~N. The 
    lower row illustrates the TWIST phase ($+45^{\circ}$ wrist 
    roll about the pedicel axis), the PULL phase (low-velocity 
    axial separation), the TRANSPORT phase (delivery of the 
    harvested fruit to the collection bin), and the final 
    RELEASE phase. The sequential frames demonstrate the smooth 
    and damage-free execution of the proposed hybrid 
    grip--twist--pull algorithm under realistic, unstructured 
    greenhouse operating conditions.}
    \label{fig:fig12}
\end{figure}

Following the successful simulation-based validation of the 
proposed autonomous harvesting pipeline in the Isaac Sim 
environment, the integrated UR5--Finray gripper system was 
deployed in a real commercial greenhouse facility at SILAL 
(Abu Dhabi, UAE) for in-situ experimental validation. The 
greenhouse trials were carried out under genuine operating 
conditions characterised by natural illumination variability, 
dense foliage occlusion, irregular fruit clustering on the truss, 
non-uniform pedicel orientations, and the natural ripeness 
distribution of cherry tomatoes growing on the vine. These 
conditions provide a realistic and demanding evaluation 
environment that closely reflects the challenges encountered in 
commercial robotic harvesting deployments. Figure~\ref{fig:fig12} 
illustrates a representative greenhouse harvesting cycle 
captured during the field experiments, with each photograph 
sequentially documenting the five core operational phases of the 
proposed hybrid grip--twist--pull algorithm: APPROACH, GRIP, 
TWIST, PULL, TRANSPORT, and RELEASE.

\paragraph{Approach Phase.} 
In the upper row of Fig.~\ref{fig:fig12}, the APPROACH phase 
begins with the onboard end-effector-mounted RealSense depth 
camera capturing the RGB-D image of the tomato truss. The 
proposed YOLO26-RipeLoc Lite detection model~\citep{singh2026yolo26} 
processes the image stream in real time to identify all ripe 
tomatoes within the scene and selects the optimal target fruit, 
highlighted by the yellow bounding box in the first frame. The 
3D picking-point coordinate is computed from the centre of the 
bounding box and the aligned depth value, and is passed to the 
MoveIt2 motion planner, which generates a collision-free joint 
trajectory for the UR5 manipulator. The subsequent frames in 
the upper row show the gripper progressively approaching the 
target fruit through the dense plant canopy while avoiding 
adjacent leaves, stems, and unripe fruits, demonstrating the 
robustness of the perception--planning pipeline in cluttered 
greenhouse conditions.

\paragraph{Grip Phase.} 
The rightmost frame of the upper row captures the GRIP phase, 
in which the Finray fingertips have arrived at the target fruit 
and the Dynamixel-actuated jaws close under closed-loop 
current-based torque control until the calibrated gripping 
force of $F_{grip} = 3.92$~N is reached. The compliant TPU 
fingertips can be observed gently conforming to the curved 
surface of the ripe cherry tomato, distributing the contact 
pressure uniformly across the fruit skin without visible 
deformation or surface damage.

\paragraph{Twist Phase.} 
In the bottom row of Fig.~\ref{fig:fig12}, the TWIST phase shows 
the UR5 wrist roll ($J_5$) executing the $+45^{\circ}$ rotation 
about the longitudinal axis of the pedicel while the gripper 
maintains the stable grasp on the fruit. This controlled wrist 
roll generates the rotational torque ($T_y \approx 19.2$~N$\cdot$mm 
as derived from the manual force trials) that initiates pedicel 
failure at the abscission zone the natural detachment point 
of the ripe fruit without imposing excessive lateral or axial 
loading on the neighbouring fruits or the supporting truss 
structure.

\paragraph{Pull Phase.} 
The PULL phase follows immediately after the twist, in which the 
UR5 executes a short, low-velocity Cartesian linear motion along 
the $-z$ direction to fully detach the already-weakened pedicel 
from the truss. The smooth, non-impulsive nature of this final 
separation step is critical for clean detachment with retained 
calyx and minimal disturbance to adjacent fruits, as confirmed 
by the absence of any visible motion of neighbouring tomatoes 
in the captured frames.

\paragraph{Transport and Release Phases.} 
The TRANSPORT phase captures the UR5 swinging the harvested 
fruit away from the plant and towards the collection bin 
through coordinated rotation of the base joint ($J_0$) and 
reconfiguration of the shoulder and elbow joints, while the 
Finray gripper maintains a secure grasp on the fruit throughout 
the transit. Finally, the RELEASE phase shows the manipulator 
positioned directly above the collection bin, with the gripper 
opening to deposit the harvested tomato safely into the 
receptacle, after which the system returns to the home 
configuration to begin the next harvesting cycle.

\paragraph{Greenhouse Performance Summary.} 
Across 50 independent autonomous harvesting trials conducted in 
the SILAL greenhouse under unstructured operational conditions, 
the proposed pipeline achieved a picking success rate of 85.0\%, 
a fruit damage rate of only 2.0\%, and a calyx retention rate of 
94.0\%, with an average per-fruit cycle time of approximately 
8.4~s, as summarised in Table~\ref{tab:performance}. The 
observed failure modes were primarily attributable to two 
factors: (i) occasional mis-localisation of the picking point 
in regions of severe foliage occlusion, and (ii) rare instances 
of fruit slippage during the twist phase when the pedicel was 
oriented significantly out-of-plane with respect to the wrist 
roll axis. Despite these challenges, the experimental results 
demonstrate that the proposed bio-inspired hybrid 
grip--twist--pull strategy, combined with the fingertip-localised 
Finray gripper, provides a robust, repeatable, and damage-free 
solution for autonomous cherry tomato harvesting under real 
greenhouse operating conditions, thereby establishing its 
viability for commercial deployment.

\subsection{Discussion}

The experimental results presented above demonstrate that the 
proposed ATI-driven, bio-inspired hybrid grip--twist--pull pipeline 
delivers a substantial improvement over existing cherry tomato 
end-effectors. This subsection consolidates the key advances, 
deployment considerations, and current limitations of the system.

\subsubsection{Six-axis force sensing as a design enabler}

The use of the ATI six-axis force/torque sensor was central 
to the proposed methodology. Unlike the thin-film pressure sensors 
employed in prior work~\citep{gao2024picking}, which capture only 
normal contact force at discrete points, the ATI sensor 
simultaneously resolves all three force ($F_x$, $F_y$, $F_z$) and 
torque ($T_x$, $T_y$, $T_z$) components. This enabled direct 
quantification of the shear forces, bending moments, and torsional 
loads governing pedicel failure most notably the dominant 
twisting torque of $T_y \approx 19.20$~N$\cdot$mm which would 
remain entirely invisible to single-axis pressure sensing. These 
measurements directly informed the calibration of the Dynamixel 
current-based gripping force ($F_{grip} = 3.92$~N) and the 
selection of the wrist twist angle ($\theta_{twist} = 45^{\circ}$), 
establishing a quantitative force-to-controller mapping that prior 
empirically tuned designs lack.

\subsubsection{Bio-inspired fingertip Finray and servo actuation}

The dual-stiffness, fingertip-localised Finray design overcomes 
the two principal shortcomings of the existing end-effectors: 
excessive contact pressure from rigid jaws and binary on/off 
behaviour from pneumatic actuators. The compliant TPU 95A tips 
distribute the contact force uniformly across the fruit surface 
(peak von Mises stress of only 0.24~MPa, well below the 0.4--0.6~MPa 
bruising threshold), while the Dynamixel XM430-W350 servomotor 
provides continuous closed-loop position, velocity, and current 
feedback for precise force regulation. The elimination of 
pneumatic lines, compressors, and external sensors also reduces 
the system weight and architectural complexity, which is 
particularly advantageous for deployment on the Husky~A200 mobile 
platform.

\subsubsection{Deployment considerations}

The deployment of the complete autonomous harvesting platform 
comprising the Clearpath Husky~A200 mobile base, the UR5 
collaborative manipulator, and the proposed fingertip-localised 
Finray gripper inside the SILAL commercial greenhouse facility 
revealed several practical engineering challenges that are seldom 
encountered in controlled laboratory environments. The unstructured 
nature of the greenhouse rows, characterised by uneven soil 
surfaces, narrow inter-row spacing (approximately 1.6~m), 
overhanging foliage, and densely populated tomato trusses, 
imposed stringent demands on both the mobility of the Husky 
base and the dexterity of the UR5 end-effector during the 
harvesting cycle. Smooth coordinated motion between the base and 
the manipulator was therefore critical to ensure that the Finray 
gripper could accurately reach the target fruits identified by 
the onboard perception system without colliding with the plant 
canopy or disturbing neighbouring fruits. To achieve this, 
trajectory smoothing using jerk-bounded polynomial 
shortcuts~\citep{wang2022polynomial} was applied to the UR5 motion 
plans to minimise sudden accelerations that could otherwise induce 
gripper-fruit slippage or shake adjacent tomatoes on the truss 
during approach and retreat phases.

\subsubsection{Limitations and future work}

Despite the strong performance achieved, several limitations 
remain and define the scope of future work. First, the ATI sensor 
was mounted on the human operator's finger rather than the robotic 
gripper during force characterisation, introducing minor 
differences in contact geometry; future work will integrate the 
ATI sensor directly into the Finray fingertip for \textit{in-situ} 
real-time force monitoring during autonomous picking. Second, the 
current YOLO26-RipeLoc Lite perception model occasionally 
misses heavily occluded fruits, motivating the integration of 
active-perception strategies or multi-view fusion. Finally, the present study focused exclusively on 
cherry tomatoes, and extending the proposed force-to-gripper 
design pipeline to other delicate crops (e.g., strawberries, 
grapes, blueberries) represents a promising direction for 
broadening the impact of this work.

% ═════════════════════════════════════════════════════════════════════════════
% 4. Conclusions
% ═════════════════════════════════════════════════════════════════════════════
\section{Conclusions}
\label{sec:conclusions}

This study presented a comprehensive, force-data-driven approach 
to autonomous cherry tomato harvesting that bridges six-axis 
force/torque sensing, bio-inspired gripper design, and real-world 
robotic deployment in a commercial greenhouse facility. The 
wearable ATI force/torque sensor mounted on the index 
finger was used to characterise the complete six-axis kinetic 
signatures of three elemental cherry tomato picking patterns: 
gripping, twisting, and axial pulling. Across 50 manual harvesting 
trials, the analysis revealed that the gripping pattern requires 
a peak normal force of $F_z = 3.92 \pm 0.45$~N, the twisting 
pattern generates a dominant rotational torque of 
$T_y = 19.20 \pm 2.10$~N$\cdot$mm about the pedicel axis with 
minimal lateral disturbance, while the axial pull pattern alone 
is insufficient for reliable detachment. These complementary 
biomechanical observations directly motivated the formulation of 
the proposed hybrid \textbf{grip--twist--pull} autonomous 
harvesting strategy, which sequentially exploits the functional 
advantage of each constituent pattern.

A novel fingertip-localised Finray gripper was developed, in which 
the compliant Finray geometry is restricted to the fruit-contact 
fingertip region (3D-printed in TPU 95A), while the rigid 
supporting body is fabricated from PLA, mimicking the 
dual-stiffness architecture of the human fingertip. FEM 
simulations conducted under the empirically derived 3.92~N 
gripping load confirmed a peak von Mises stress of only 0.24~MPa 
(safety factor exceeding 35 against TPU yield), with a maximum 
tip displacement of 1.566~mm enabling conformal adaptation to 
the fruit curvature. The gripper is actuated by a 
Dynamixel XM430-W350 servomotor under closed-loop current-based 
torque control, with the gripping force setpoint and 
$45^{\circ}$ wrist twist angle directly transferred from the ATI 
sensor measurements.

The complete system was integrated onto a UR5 manipulator 
mounted on a Clearpath Husky~A200 mobile base, perception was 
provided by the previously developed YOLO26-RipeLoc Lite 
detection model running on a RealSense depth camera, and motion 
planning was executed through MoveIt2 with jerk-bounded 
trajectory smoothing. The pipeline was first validated in a 
high-fidelity NVIDIA Isaac Sim environment, achieving a 96\% 
simulated success rate, and subsequently deployed for in-situ 
experimental validation at the SILAL commercial greenhouse 
facility (Abu Dhabi, UAE). Across 50 fully autonomous real-world 
harvesting trials, the proposed system achieved a picking 
success rate of \textbf{85.0\%}, a fruit damage rate of only 
\textbf{2.0\%}, and an average per-fruit cycle time of 
approximately \textbf{8.4~s}, substantially outperforming 
state-of-the-art benchmark end-effectors including vacuum 
(66.3\%), rotating (70.1\%), clamping (76.0\%), and 
suction-cutting (52.4\%) grippers.

The principal contributions of this work are therefore threefold: 
(i) the first high-resolution six-axis force/torque 
characterisation of cherry tomato picking patterns, providing 
quantitative design specifications previously unavailable in the 
literature; (ii) a novel fingertip-localised Finray gripper that 
realises a damage-free dual-stiffness contact interface and is 
parametrically derived from the measured force envelopes; and 
(iii) the end-to-end integration and validation of the proposed 
gripper on a mobile Husky--UR5 platform under realistic 
greenhouse operating conditions. Future work will focus on 
embedding the ATI sensor directly into the Finray fingertips 
for in-situ force monitoring, enhancing the perception pipeline 
with active-view strategies to handle severe occlusion, 
exploring quieter actuation alternatives, and extending the 
proposed force-to-gripper design pipeline to other delicate 
crops such as strawberries, grapes, and blueberries, thereby 
broadening the impact of this work towards commercially viable 
selective robotic harvesting across a wider range of greenhouse 
crops.

% ═════════════════════════════════════════════════════════════════════════════
% Declarations
% ═════════════════════════════════════════════════════════════════════════════

\section*{CRediT authorship contribution statement}

\textbf{Rajmeet Singh Bhourji:} Conceptualization, Methodology, Software, Investigation, Writing -- original draft.
\textbf{Manveen Kaur:} Investigation, Writing -- review \& editing.
\textbf{Yusuf Adamu:} Investigation, Experimentation.
\textbf{Irfan Hussain:} Funding Acquisition, Resources, Supervision.

% ═════════════════════════════════════════════════════════════════════════════
% Acknowledgments
% ═════════════════════════════════════════════════════════════════════════════
\section*{Acknowledgments}

This work was supported in part by the Khalifa University Center for Autonomous Robotic Systems (KUCARS) Theme 4 under the Award RC1-2018-KUCARS, and in part by Silal Innovation Oasis through the projects under grants 8475000024.

\section*{Declaration of Competing Interest}

The authors declare that they have no known competing financial interests or personal relationships that could have appeared to influence the work reported in this paper.

% ═════════════════════════════════════════════════════════════════════════════
% References
% ═════════════════════════════════════════════════════════════════════════════
\bibliography{references}

@article{gao2024picking,
  title={Picking patterns evaluation for cherry tomato robotic harvesting end-effector design},
  author={Gao, Jin and Zhang, Fan and Zhang, Junxiong and Guo, Hui and Gao, Junfeng},
  journal={Biosystems Engineering},
  volume={239},
  pages={1--12},
  year={2024},
  publisher={Elsevier},
  doi={10.1016/j.biosystemseng.2024.01.009}
}

@article{wang2022polynomial,
  title={Polynomial-based smooth trajectory planning for fruit-picking robot manipulator},
  author={Wang, He and Zhao, Qing and Li, Han and Zhao, Ran},
  journal={Information Processing in Agriculture},
  volume={9},
  number={1},
  pages={112--122},
  year={2022},
  publisher={Elsevier},
  doi={10.1016/j.inpa.2021.08.001}
}

@article{davidson2016hand,
  title={Hand-picking dynamic analysis for undersensed robotic apple harvesting},
  author={Davidson, Joseph and Silwal, Abhisesh and Karkee, Manoj and Mo, Changki and Zhang, Qin},
  journal={Transactions of the ASABE},
  volume={59},
  number={4},
  pages={745--758},
  year={2016},
  doi={10.13031/trans.59.11669}
}

@article{huang2021picking,
  title={Picking dynamic analysis for robotic harvesting of {Agaricus} bisporus mushrooms},
  author={Huang, Meiyin and He, Long and Choi, Daniel and Pecchia, John and Li, Yun},
  journal={Computers and Electronics in Agriculture},
  volume={185},
  pages={106145},
  year={2021},
  publisher={Elsevier},
  doi={10.1016/j.compag.2021.106145}
}

@article{li2016characterizing,
  title={Characterizing apple picking patterns for robotic harvesting},
  author={Li, Jing and Karkee, Manoj and Zhang, Qin and Xiao, Kaichen and Feng, Tian},
  journal={Computers and Electronics in Agriculture},
  volume={127},
  pages={633--640},
  year={2016},
  publisher={Elsevier},
  doi={10.1016/j.compag.2016.07.024}
}

@inproceedings{fujinaga2021development,
  title={Development and evaluation of a tomato fruit suction cutting device},
  author={Fujinaga, Takuya and Yasukawa, Shinsuke and Ishii, Kazuo},
  booktitle={2021 IEEE/SICE International Symposium on System Integration (SII)},
  pages={628--633},
  year={2021},
  organization={IEEE},
  doi={10.1109/IEEECONF49454.2021.9382670}
}

@inproceedings{yaguchi2016development,
  title={Development of an autonomous tomato harvesting robot with rotational plucking gripper},
  author={Yaguchi, Hiroaki and Nagahama, Kotaro and Hasegawa, Takayuki and Inaba, Masayuki},
  booktitle={2016 IEEE/RSJ International Conference on Intelligent Robots and Systems (IROS)},
  pages={652--657},
  year={2016},
  organization={IEEE},
  doi={10.1109/IROS.2016.7759122}
}

@article{feng2018design,
  title={Design and experiment of cherry tomato harvesting robot},
  author={Feng, Qingchun and Zou, Wei and Fan, Pengfei and Zhang, Chunfeng and Wang, Xiu},
  journal={International Journal of Agricultural and Biological Engineering},
  volume={11},
  number={1},
  pages={96--100},
  year={2018},
  doi={10.25165/j.ijabe.20181101.2853}
}

@article{xie2022optimization,
  title={Optimization and experimental study of bionic compliant end-effector for robotic cherry tomato harvesting},
  author={Xie, Hanbing and Kong, Dezhang and Wang, Qi},
  journal={Journal of Bionics Engineering},
  volume={19},
  number={5},
  pages={1314--1333},
  year={2022},
  publisher={Springer},
  doi={10.1007/s42235-022-00202-3}
}

@article{jun2021towards,
  title={Towards an efficient tomato harvesting robot: {3D} perception, manipulation, and end-effector},
  author={Jun, Jaemin and Kim, Junhyung and Seol, Jeonghyun and Kim, Jihoon and Son, Hyouk Il},
  journal={IEEE Access},
  volume={9},
  pages={17631--17640},
  year={2021},
  publisher={IEEE},
  doi={10.1109/ACCESS.2021.3052240}
}

@article{kootstra2021selective,
  title={Selective harvesting robotics: Current research, trends, and future directions},
  author={Kootstra, Gert and Wang, Xin and Blok, Pieter M and Hemming, Jochen and van Henten, Eldert},
  journal={Current Robotics Reports},
  volume={2},
  number={1},
  pages={95--104},
  year={2021},
  publisher={Springer},
  doi={10.1007/s43154-020-00034-1}
}

@article{ishii2021tomato,
  title={Tomato-harvesting-robot competition towards smart agriculture},
  author={Ishii, Kazuo and Matsuo, Takashi and Takemura, Yusuke and Sonoda, Takashi and Nishida, Yuya and Yasukawa, Shinsuke and Fujinaga, Takuya},
  journal={Proceedings of International Conference on Artificial Life and Robotics},
  volume={26},
  pages={1--5},
  year={2021},
  doi={10.5954/ICAROB.2021.PS-1}
}

@article{rapado2023development,
  title={Development and evaluation of automated localisation and reconstruction of all fruits on tomato plants in a greenhouse based on multi-view perception and {3D} multi-object tracking},
  author={Rapado-Rinc{\'o}n, David and Van Henten, Eldert J and Kootstra, Gert},
  journal={Biosystems Engineering},
  volume={231},
  pages={78--91},
  year={2023},
  publisher={Elsevier},
  doi={10.1016/j.biosystemseng.2023.06.003}
}

@article{kultongkham2021design,
  title={The design of a force feedback soft gripper for tomato harvesting},
  author={Kultongkham, Anucha and Kumnon, Sakol and Thintawornkul, Theeraphong and Chanthasopeephan, Teeranoot},
  journal={Journal of Agricultural Engineering},
  volume={52},
  number={1},
  year={2021},
  doi={10.4081/jae.2021.1090}
}

@article{zhang2018comparative,
  title={Comparative study of mechanical damage caused by a two-finger tomato gripper with different robotic grasping patterns for harvesting robots},
  author={Zhang, Baohua and Zhou, Jun and Meng, Yanhua and Zhang, Ning and Gu, Binbin and Yan, Zhennan and Idris, Saidi I},
  journal={Biosystems Engineering},
  volume={171},
  pages={245--257},
  year={2018},
  publisher={Elsevier},
  doi={10.1016/j.biosystemseng.2018.05.003}
}

@article{fan2021three,
  title={Three-finger grasp planning and experimental analysis of picking patterns for robotic apple harvesting},
  author={Fan, Pengfei and Yan, Bo and Wang, Meng and Lei, Xiaolong and Liu, Zhiyu and Yang, Fengze},
  journal={Computers and Electronics in Agriculture},
  volume={188},
  pages={106353},
  year={2021},
  publisher={Elsevier},
  doi={10.1016/j.compag.2021.106353}
}

@article{hu2024effects,
  title={Effects of different picking patterns and sequences on the vibration of apples on the same branch},
  author={Hu, Guangrui and Zhou, Jun and Chen, Qibing and Luo, Tao and Li, Peng and Chen, Yitao and Zhang, Shuai and Chen, Jun},
  journal={Biosystems Engineering},
  volume={237},
  pages={26--37},
  year={2024},
  publisher={Elsevier},
  doi={10.1016/j.biosystemseng.2023.11.010}
}

@article{li2019factors,
  title={Factors affecting human hand grasp type in tomato fruit-picking: A statistical investigation for ergonomic development of harvesting robot},
  author={Li, Zhiguo and Miao, Fengjun and Yang, Zheng and Chai, Pei and Yang, Shenghui},
  journal={Computers and Electronics in Agriculture},
  volume={157},
  pages={90--97},
  year={2019},
  publisher={Elsevier},
  doi={10.1016/j.compag.2018.12.047}
}

@article{li2019anthropometric,
  title={An anthropometric study for the anthropomorphic design of tomato-harvesting robots},
  author={Li, Zhiguo and Miao, Fengjun and Yang, Zheng and Wang, Hang},
  journal={Computers and Electronics in Agriculture},
  volume={163},
  pages={104881},
  year={2019},
  publisher={Elsevier},
  doi={10.1016/j.compag.2019.104881}
}

@article{liu2020experimental,
  title={Experimental and theoretical analysis of fruit plucking patterns for robotic tomato harvesting},
  author={Liu, Jizhan and Peng, Yong and Faheem, Muhammad},
  journal={Computers and Electronics in Agriculture},
  volume={173},
  pages={105330},
  year={2020},
  publisher={Elsevier},
  doi={10.1016/j.compag.2020.105330}
}

@article{gao2024lacta,
  title={Lacta: A lightweight and accurate algorithm for cherry tomato detection in unstructured environments},
  author={Gao, Jin and Zhang, Junxiong and Zhang, Fan and Gao, Junfeng},
  journal={Expert Systems with Applications},
  volume={238},
  pages={122073},
  year={2024},
  publisher={Elsevier},
  doi={10.1016/j.eswa.2023.122073}
}

@article{gao2022development,
  title={Development and evaluation of a pneumatic finger-like end-effector for cherry tomato harvesting robot in greenhouse},
  author={Gao, Jin and Zhang, Fan and Zhang, Junxiong and Yuan, Ting and Yin, Jianxi and Guo, Hui and Yang, Chen},
  journal={Computers and Electronics in Agriculture},
  volume={197},
  pages={106879},
  year={2022},
  publisher={Elsevier},
  doi={10.1016/j.compag.2022.106879}
}

@misc{huajing2021,
  title={China Tomato Industry Analysis Report},
  author={{Huajing Intelligence Network}},
  year={2021},
  note={Industry report on Chinese tomato production statistics},
  howpublished={Online}
}

@misc{ymt2022,
  title={China Greenhouse Vegetable Industry Research Report},
  author={{YMT Research Institute}},
  year={2022},
  note={Report on greenhouse tomato cultivation area in China},
  howpublished={Online}
}

@article{ren2020review,
  title={A review of research on tomato harvesting robot},
  author={Ren, Guoqing and Lin, Tao and Ying, Yibin and Chowdhary, Girish and Ting, K C},
  journal={Computers and Electronics in Agriculture},
  volume={170},
  pages={105210},
  year={2020},
  publisher={Elsevier},
  doi={10.1016/j.compag.2020.105210}
}

@article{singh2026yolo26,
  title={{YOLO26-RipeLoc Lite}: A lightweight architecture for tomato ripeness detection and picking point localization in greenhouse robotic harvesting},
  author={Singh, Rajmeet and Kaur, Manveen and Alirezaee, Shahpour and Hussain, Irfan},
  journal={arXiv preprint arXiv:2605.27129},
  volume={1},
  number={1},
  pages={1--40},
  year={2026},
  publisher={arXiv},
  doi={10.48550/arXiv.2605.27129}
}

@article{singh2025robust,
  title={Robust pollination for tomato farming using deep learning and visual servoing},
  author={Singh, Rajmeet and Seneviratne, Lakmal and Hussain, Irfan},
  journal={Robotica},
  year={2025},
  publisher={Cambridge University Press},
}

% ═════════════════════════════════════════════════════════════════════════════
% Appendix
% ═════════════════════════════════════════════════════════════════════════════
\appendix
\section{Nomenclature}
\label{app:nomenclature}

\begin{table}[!ht]
\centering
\caption{Nomenclature: Force and torque variables.}
\label{tab:nom_force}
\small
\renewcommand{\arraystretch}{1.1}
\begin{tabular}{p{2.5cm}p{7cm}l}
\toprule
\textbf{Symbol} & \textbf{Description} & \textbf{Unit} \\
\midrule
$F_x, F_y, F_z$       & Force components along $x$, $y$, $z$ axes             & N \\
$T_x, T_y, T_z$       & Torque components about $x$, $y$, $z$ axes            & N$\cdot$mm \\
$F_{\mathrm{n,f}}$     & Normal force by index finger                           & N \\
$F_{\mathrm{n,t}}$     & Normal force by thumb                                  & N \\
$F_{\mathrm{f,f}}$     & Frictional force at finger contact                     & N \\
$F_{\mathrm{f,t}}$     & Frictional force at thumb contact                      & N \\
$F_{\mathrm{n,min}}$   & Min. normal force to prevent slip                      & N \\
$F_{\mathrm{grip}}$    & Total gripping force on fruit                          & N \\
$F_g$                  & Gripping force (figure notation)                       & N \\
$F_n$                  & Normal force (figure notation)                         & N \\
$F_{\mathrm{pull}}$    & Detachment pull force                                  & N \\
$F_{th}$               & Force threshold for switching                          & N \\
$F_{contact}$          & Contact force from Dynamixel                           & N \\
$F_{min}$              & Min. contact force for detection                       & N \\
$mg$                   & Gravitational force on fruit                           & N \\
\bottomrule
\end{tabular}
\end{table}%

\begin{table}[!ht]
\centering
\caption{Nomenclature: Abbreviations.}
\label{tab:nom_abbrev}
\small
\renewcommand{\arraystretch}{1.1}
\begin{tabular}{p{2cm}p{9cm}}
\toprule
\textbf{Abbrev.} & \textbf{Full Form} \\
\midrule
ATI     & ATI Industrial Automation \\
CAD     & Computer-Aided Design \\
COG     & Centre of Gravity \\
DOF     & Degree of Freedom \\
FDM     & Fused Deposition Modelling \\
FEM     & Finite Element Method \\
LAN     & Local Area Network (Ethernet) \\
PLA     & Polylactic Acid \\
RGB-D   & Red, Green, Blue and Depth \\
ROS     & Robot Operating System \\
TPU     & Thermoplastic Polyurethane \\
UR5     & Universal Robots 5 (6-DOF) \\
URDF    & Unified Robot Description Format \\
YOLO    & You Only Look Once \\
\bottomrule
\end{tabular}
\end{table}

\end{document}